%% file: ALuMA.tex
\documentclass{article} % For LaTeX2e
\usepackage{jmlr2e,times,natbib}
\usepackage{wrapfig}
\input{header.tex}

\usepackage{tikz}
\usetikzlibrary{arrows}
\usetikzlibrary{calc}

\newcommand{\Vol}{\mathrm{Vol}}
\newcommand{\VolEst}{\mathrm{VolEst}}
\newcommand{\ball}{\mathbb{B}_1}
\newcommand{\avg}{\mathrm{avg}}
\newcommand{\alg}{\mathrm{alg}}
\newcommand{\thr}{\mathrm{line}}
\newcommand{\cW}{\mathcal{W}}
\newcommand{\determined}{\Lleftarrow}
\newcommand{\avgCost}{c_{\mathrm{avg}}}
\newcommand{\wCost}{c_{\mathrm{wc}}}
\newcommand{\ceil}[1]{\lceil{#1}\rceil}

\newcommand{\OPT}{\ensuremath{\mathrm{OPT}}}
\newcommand{\algref}[1]{Alg.~\ref{#1}}
\newcommand{\favg}{f_{\mathrm{avg}}}
\newcommand{\cA}{\mathcal{A}}

\newcommand{\fix}[1]{#1}

\jmlrheading{}{}{}{}{}{Gonen, Sabato and Shalev-Shwartz}

\ShortHeadings{Efficient Active Learning of Halfspaces}{Gonen, Sabato and Shalev-Shwartz}
\firstpageno{1}

\title{Efficient Active Learning of Halfspaces: an Aggressive Approach}

\author{\name{Alon Gonen} \email{alongnn@cs.huji.ac.il}\\
 \addr Benin school of Computer Science and Engineering \\
The Hebrew University\\
Givat Ram, Jerusalem 91904, Israel\\
\AND
\name{Sivan Sabato} \email{sivan.sabato@microsoft.com}\\
\addr Microsoft Research New England\\
1 Memorial Drive\\
Cambridge, MA, 02142\\
\AND
\name{Shai Shalev-Shwartz} \email{shais@cs.huji.ac.il}\\
 \addr Benin school of Computer Science and Engineering \\
The Hebrew University\\
Givat Ram, Jerusalem 91904, Israel
 }

\editor{}

\begin{document}

\maketitle

\begin{abstract}
We study pool-based active learning of half-spaces. 
We revisit the aggressive approach for active learning in the realizable case,
and show that it can be made efficient and
practical, while also having theoretical guarantees under reasonable
assumptions. We further show, both theoretically and experimentally,
that it can be preferable to mellow approaches.  Our efficient aggressive active learner of half-spaces has formal approximation guarantees that hold when
the pool is separable with a margin. While our analysis is focused on
the realizable setting, we show that a simple heuristic allows using
the same algorithm successfully for pools with low error as well. 
We further compare the aggressive
approach to the mellow approach, and prove that there are cases in
which the aggressive approach results in
significantly better label complexity compared to the mellow approach.
We demonstrate experimentally that substantial improvements in
label complexity can be achieved using the aggressive approach, for both
realizable and low-error settings.\footnote{A short version of this paper was accepted to ICML 2013.}

\end{abstract}

\section{Introduction}  \label{sec:intro}

%Pool based active learning
We consider pool-based active learning \citep{McCallum}, in
which a learner receives a pool of unlabeled examples, and can iteratively
query a teacher for the labels of examples from the pool. The goal of the
learner is to return a low-error prediction rule for the labels of the examples, using a small number of queries.  The number of queries used by the learner is termed its \emph{label
  complexity}. This setting is most useful when unlabeled data is abundant but
labeling is expensive, a common case in many data-laden applications.
A pool-based algorithm can be
used to learn a classifier in the standard PAC model, while
querying fewer labels. This can be done by first drawing a random unlabeled sample to be used
as the pool, then using pool-based active learning to identify its labels with few queries,
and then using the resulting labeled sample as input to a regular ``passive'' PAC-learner.

%mellow and agressive algorithms
Most active learning approaches can be loosely described as more `aggressive' or more `mellow'. A more aggressive approach is one in which only highly informative queries are requested (where the meaning of `highly informative' depends on the particular algorithm) \citep{TongKoller,Balcan07, Kalai}, while the mellow approach, first proposed in the CAL algorithm \citep{CAL}, is one in which the learner essentially queries all the labels it has not inferred yet. 

In recent years a significant advancement has been made for active learning in the PAC model.
In particular, it has been shown that when the data is realizable (relative to some assumed hypothesis class), the mellow approach can guarantee an exponential improvement in label complexity, compared to passive learning \citep{A2}.
This exponential improvement depends on the properties of the
distribution, as quantified by the \emph{Disagreement Coefficient}
proposed in \citet{Hanneke07a}. Specifically, when learning
half-spaces in Euclidean space, the disagreement coefficient implies a
low label complexity when the data distribution is uniform or close to
uniform. Guarantees have also been shown for the case where the data distribution is a finite mixture of Gaussians \citep{el2012active}.

An advantage of the mellow approach is its ability to obtain label complexity improvements in the agnostic setting, which allows an arbitrary and large labeling error \citep{A2, dasgupta2007general}. Nonetheless, in the realizable case the mellow approach is not always optimal, even for the uniform distribution \citep{Balcan07}. In this work we revisit the aggressive approach for the realizable case, and in particular for active learning of half-spaces in Euclidean space. We show that it can be made efficient and practical, while also having theoretical guarantees under reasonable assumptions. We further show, both theoretically and experimentally, that it can sometimes be preferable to mellow approaches.

In the first part of this work we construct an efficient aggressive
active learner for half-spaces in Euclidean space, which is
approximately optimal, i.e. achieves near-optimal label complexity, if the pool is separable with a margin. While our analysis is focused on the realizable setting, we show that a simple heuristic allows using the same algorithm successfully for pools with low error as well. Our algorithm for halfspaces is based on a greedy query selection approach as proposed in \citet{TongKoller,Dasgupta05}. We obtain improved target-dependent approximation guarantees for greedy selection in a general active learning setting. These guarantees allow us to prove meaningful approximation guarantees for halfspaces based on a margin assumption.

In the second part of this work we compare the greedy approach to the mellow approach. 
We prove that there are cases in which this highly aggressive greedy approach results in significantly better label complexity compared to the mellow approach.
We further demonstrate experimentally that substantial improvements in label complexity can be achieved compared to mellow approaches, for both realizable and low-error settings.

The first greedy query selection algorithm for learning halfspaces in Euclidean space was proposed by \citet{TongKoller}. The greedy algorithm is based on the notion of a \emph{version space}: the set of all hypotheses in the hypothesis class that are consistent with the labels currently known to the learner. In the case of halfspaces, each version space is a convex body in Euclidean space. Each possible query thus splits the current version space into two parts: the version space that would result if the query received a positive label, and the one resulting from a negative label. Tong and Koller proposed to query the example from the pool that splits the version space
as evenly as possible. To implement this policy, one would need to calculate the volume of a convex body in Euclidean space, a problem which is known to be computationally intractable \citep{Brightwell}.  Tong and Koller thus
implemented several heuristics that attempt to follow their proposed
selection principle using an efficient algorithm.  For instance, they
suggest to choose the example which is closest to the max-margin
solution of the data labeled so far. However, none of their heuristics
provably follow this greedy selection policy. 

The label complexity of greedy pool-based active learning algorithms can be analyzed
by comparing it to the best possible label complexity of any pool-based active learner
on the same pool. The \emph{worst-case label complexity} of an active learner
is the maximal number of queries it would make on the given pool, where the maximum
is over all the possible classification rules that can be consistent with the pool
according to the given hypothesis class. The \emph{average-case label complexity}
of an active learner is the average number of queries it would make on the given pool,
where the average is taken with respect to some fixed probability distribution $P$
over the possible classifiers in the hypothesis class. For each of these definitions,
the optimal label complexity is the lowest label complexity that can be achieved by 
an active learner on the given pool. Since implementing the optimal label complexity
is usually computationally intractable, an alternative is to implement an efficient algorithm, and to guarantee a bounded factor of approximation on its label complexity, compared to the optimal
label complexity.

\citet{Dasgupta05} showed that if a greedy algorithm splits the probability
mass of the version space as evenly as possible, as defined by the fixed probability distribution $P$ over the hypothesis class, then the approximation factor for its average label complexity, with respect to the same distribution, is bounded by $O(\log(1/p_{\min}))$, where $p_{\min}$ is the minimal probability of any possible labeling of the pool, if the classifier is drawn according to the fixed distribution.
\citet{Golovin} extended Dasgupta's result and showed that a
similar bound holds for an approximate greedy rule. They 
also showed that the approximation factor for the worst-case label complexity of an approximate
greedy rule is also bounded by $O(\log(1/p_{\min}))$, thus extending a result of \citet{Arkin}. 
Note that in the worst-case analysis, the fixed distribution is only
an analysis tool, and does not represent any assumption on the true
probability of the possible labelings. 

Returning to greedy selection of halfspaces in Euclidean space, 
we can see that the fixed distribution over hypotheses that matches the volume-splitting strategy
is the distribution that draws a halfspace uniformly from the unit ball.\footnote{We discuss the challenges presented by
  other natural choices of a distribution in \secref{sec:difficulty}} 
The analysis presented above thus can result in poor approximation factors, since if there are instances in the pool that are very close to each other, then $p_{\min}$ might be
very small. 

We first show that mild conditions suffice to
guarantee that $p_{\min}$ is bounded from below. By proving a
variant of a result due to \citet{MurogaToTa61}, we show that if the examples in
the pool are stored using number of a finite accuracy $1/c$, then
$p_{\min} \geq (c/d)^{d^2}$, where $d$ is the dimensionality of the space. It follows that the approximation factor for the worst-case label complexity of our algorithm is at most $O(d^2\log(d/c))$. 

While this result provides us with a uniform lower bound on
$p_{\min}$, in many real-world situations
the probability of the target hypothesis (i.e., one that is consistent with the true labeling) could be much larger than $p_{\min}$. A noteworthy
example is when the target hypothesis separates the pool with a
margin of $\gamma$. In this case, it can be shown that the probability
of the target hypothesis is at least $\gamma^d$, which can
be significantly larger than $p_{\min}$. An immediate question is therefore:
can we obtain a \emph{target-dependent} label complexity approximation factor
that would depend on the probability of the target hypothesis, $P(h)$, instead of the minimal probability of any labeling?

We prove that such a target dependent bound \emph{does not} hold for
a general approximate-greedy algorithm. To overcome this,
we introduce an algorithmic change to the approximate greedy policy, which allows
us to obtain a label complexity approximation factor of $\log(1/P(h))$.
 This can be achieved by running the approximate-greedy procedure,
but stopping the procedure early, before reaching a pure version space that exactly matches
 the labeling of the pool. Then, an approximate
majority vote over the version space, that is,  a random rule which approximates the majority vote
with high probability, can be used to determine the labels of the pool.
This result is general and holds for any hypothesis class and distribution.
For halfspaces, it implies an approximation-factor guarantee of $O(d\log(1/\gamma))$. 

We use this result to provide an efficient approximately-optimal active learner for half-spaces, called \emph{ALuMA}, which relies on randomized approximation
of the volume of the version space \citep{Lovasz}. This allows us to prove a margin-dependent approximation factor guarantee for ALuMA. We further show an additional, more practical implementation of the algorithm, which has similar guarantees under mild conditions which often hold in practice. The assumption of separation with a margin can be relaxed if a lower
bound on the total hinge-loss of the best separator for the pool can
be assumed. We show that under such an assumption a simple
transformation on the data allows running ALuMA as if the data was
separable with a margin. This results in approximately optimal label
complexity with respect to the new representation.

We also derive lower bounds, showing that the dependence of our
label-complexity guarantee on the accuracy $c$, or the margin
parameter $\gamma$, is indeed necessary and is not an artifact of our
analysis. We do not know if the dependence of our bounds on $d$ is
tight. It should be noted that some of the most popular learning algorithms (e.g. SVM, Perceptron, and
AdaBoost) rely on a large-margin assumption to derive dimension-independent
sample complexity guarantees. In contrast, here we use the margin for computational reasons. Our approximation guarantee depends
logarithmically on the margin parameter, while the sample complexities of SVM,
Perceptron, and AdaBoost depend polynomially on the margin. Hence, we require
a much smaller margin than these algorithms do.
In a related work, \citet{Balcan07} proposed an active learning algorithm with dimension-independent guarantees under a margin assumption. These guarantees hold for a restricted class of data distributions.

In the second part of this work, we compare the greedy approach to the mellow approach of CAL in the realizable case, both theoretically and experimentally. Our theoretical results show the following:
\begin{enumerate}
\item In the simple learning setting of thresholds on the line, our margin-based approach is preferable to the mellow approach when the true margin of the target hypothesis is large.
\item There exists a distribution in Euclidean space such that the mellow approach cannot achieve a significant improvement in label complexity over passive learning for halfspaces, while the greedy approach achieves such an improvement using more unlabeled examples.
\item There exists a pool in Euclidean space such that the mellow approach requires exponentially more labels than the greedy approach.
\end{enumerate}
We further compare the two approaches experimentally, both on separable data and on data with small error. The empirical evaluation indicates that our algorithm, which can be implemented in practice, achieves state-of-the-art results. It further suggests that aggressive approaches
can be significantly better than mellow approaches in some practical settings.

\section{On the challenges in active learning for halfspaces}\label{sec:difficulty}

The approach we employ for active learning does not provide
absolute guarantees for the label complexity of learning, but a relative
guarantee instead, in comparison with the optimal label complexity. 
One might hope that an absolute guarantee could be achieved using a different algorithm, for instance in the case of half-spaces.
However, the following example from \cite{Dasgupta05} indicates that no meaningful guarantee
can be provided that holds for all possible pools. 

% margin alone is not enough

\begin{ex}\label{ex:dasgupta}  
Consider a distribution in $\reals^d$ for any $d \geq 3$. 
Suppose that the support of the
distribution is a set of evenly-distributed points on a two-dimensional sphere that does not
circumscribe the origin, as illustrated in the following figure. As
can be seen, each point can be separated from the rest of the points with
a halfspace. 
\end{ex}
\begin{center}
\begin{tikzpicture}
%\draw (0,0) circle (1); 
\foreach \angle in {0,40,...,320} 
 \fill (\angle:1) circle (0.05);
%\draw (0.9,1) -- (0.9,-1);
\draw[dashed] (1.4,-0.2) -- (0,1.3);
\end{tikzpicture}
\end{center}
In this example, to distinguish between the case in which
all points have a negative label and the case in which one of the points has a positive label while the rest have a negative label, any active learning algorithm will have to query every point at least once. It follows that for any $\epsilon>0$, if the number
of points is $1/\epsilon$, then the label complexity to achieve an
error of at most $\epsilon$ is $1/\epsilon$. On the other hand, the sample
complexity of passive learning in this case is order of
$\frac{1}{\epsilon} \log \frac{1}{\epsilon}$, hence no active
learner can be significantly better than a passive learner on this
distribution. 

Since we provide margin-dependent guarantees, one may wonder if a margin assumption alone can
guarantee that few queries suffice to learn the half-space. This is not the case, as evident by the following variation of \exref{ex:dasgupta}.

 \begin{ex}\label{ex:margin}
   Let $\gamma \in (0,\half)$ be a margin parameter. Consider a pool of $m$
   points in $\reals^d$, such that all the points are on the unit
   sphere, and for each pair of points $x_1$ and $x_2$,
   $\inner{x_1,x_2} \leq 1-2\gamma$. It was shown in \citep{shannon59}
   that for any $m \leq O(1/\gamma^d)$, there exists a set of points
   that satisfy the conditions above.  For any point $x$ in such a
   pool, there exists a (biased) halfspace that separates $x$ from
   the rest of the points with a margin of $\gamma$. 
  This can be seen
   by letting $w = x$ and $b = 1-\gamma$.  Then $\inner{w,x} - b =
   \gamma$ while for any $z\neq x$ in the set, $\inner{w,z} - b =
   \inner{x,z} - 1 + \gamma \leq -\gamma$. 
   By adding a single
   dimension, this example can be transformed to one with
   homogeneous (unbiased) halfspaces. Each point in this pool can be separated from the rest of the points by a halfspace. Thus, if the correct labeling is all-positive, then all $m$ examples need to be queried to label the pool correctly. 
 \end{ex}

These examples show that there are ``difficult'' pools, where no active learner can do well. 
The advantage of the greedy approach is that the optimal label complexity is used as a natural measure of the difficulty of the pool.

At first glance it might seem that there are simpler ways to implement
an efficient greedy strategy for halfspaces, by using a different distribution over the hypotheses.  For instance, if there are $m$ examples in $d$ dimensions, Sauer's lemma states that
the effective size of the hypothesis class of halfspaces will be at most $m^d$. One can thus use the uniform distribution over this finite class, and greedily reduce the number of possible hypotheses in the version space, obtaining a $d\log(m)$ factor relative to the optimal label complexity. However, a direct implementation of this method will be exponential in $d$, and it is not clear whether this approach has a polynomial implementation.

 % Why simple discretization does not work

 Another approach is to discretize the version space, by considering
 only halfspaces that can be represented as vectors on a $d$-dimensional grid
 $\{-1,-1+c,\ldots,1-c,1\}^d$.  This results in a finite hypothesis class of
 size $\left(2/c+1\right)^d$, and we get an approximation factor of $O(d \log(1/c))$ for the
 greedy algorithm, compared to an optimal algorithm on the same finite
 class. However, it is unknown whether a greedy algorithm for reducing the
 number of such vectors in a version space can be implemented efficiently, 
 since even determining whether a single grid point exists in
 a given version space is NP-hard \citep[see e.g.][Section 2.2]{Matousek}.
 In particular, the volume of the version space cannot
 be used to estimate this quantity, since the volume of a body and the number of grid points in this body are not correlated. For example, consider a line in $\reals^2$, whose
   volume is $0$. It can contain zero grid points or many grid points, depending on its alignment with respect to the grid.
     Therefore,
 the discretization approach is not straightforward as one might first assume. In fact, if this approach is at all computationally
 feasible, it would probably require the use of some approximation
 scheme, similarly to the volume-estimation approach that we describe below.

% reducing the number of uncertain labels.

Yet another possible direction for pool-based active learning is to greedily select a query whose answer would determine the labels of the largest amount of pool examples. The main challenge in this
direction is how to analyze the label complexity of such an algorithm: it is
unclear whether competitiveness with the optimal label complexity can be guaranteed in this
case. Investigating this idea, both theoretically and experimentally, is an
important topic for future work.  Note that the CAL algorithm \citep{CAL}, which we
discuss in \secref{sec:comparison}, can be seen as implementing a mellow version of
this approach, since it decreases the so-called ``disagreement region'' in
each iteration.

\section{Definitions and Preliminaries}\label{sec:def}

In pool-based active learning, the learner receives as input a set of
instances, denoted $X = \{x_1,\ldots,x_m\}$.  Each instance $x_i$ is
associated with a label $L(i) \in \{\pm 1\}$, which is initially unknown to the
learner. The learner has access to a teacher, represented by the
oracle $L:[m] \rightarrow \{-1,1\}$. \fix{An active learning algorithm $\cA$ obtains $(X,L,T)$ as input,
  where $T$ is an integer which represents the label budget of $\cA$}. The goal of the learner is to
find the values $L(1),\ldots,L(m)$ using as few calls to $L$ as
possible. We assume that $L$ is determined by a function $h$ taken
from a predefined hypothesis class $\cH$. Formally, for an oracle $L$ and a hypothesis $h \in \cH$, we write $L \determined h$ to state that for all $i$, $L(i)=h(x_i)$.

Given $S \subseteq X$ and $h \in \H$, we denote the partial realization of $h$
on $S$ by
\begin{equation}  \label{eq:partial}
h|_S = \{(x,h(x)) : x \in S\} ~.
\end{equation}
We denote by $V(h|_S)$ the
version space consisting of the hypotheses which are consistent with
$h|_S$. Formally,
\[
V(h|_S) = \{h' \in \H: \forall x \in S,~~h'(x) = h(x)~\}.
\]

\fix{Given $X$ and $\cH$, we define, for each $h \in \cH$, the equivalence class of $h$ over $\cH$, $[h] = \{ h' \in \cH \mid \forall x \in X,\, h(x) =
  h'(x) \}$. We consider a probability distribution $P$ over $\cH$ such that $P([h])$ is defined for all $h\in \cH$. For brevity, we denote $P(h) = P([h])$. Similarly, for a set $V \subseteq \cH$, $P(V) = P(\cup_{h\in V}[h])$. Let $p_{\min} = \min_{h \in \cH} P(h)$. } 

We specifically consider the hypothesis class of homogeneous halfspaces in
$\reals^d$. In this case, $X \subseteq \reals^d$. The hypothesis class $\H$ is defined by $\cW = \{ x \mapsto \sgn(\inner{w,x}) \mid w \in \reals^d \}$, where $\inner{w,x}$ is the
inner product between the vectors $w$ and $x$. 

For a given active learning algorithm $\cA$, we denote by $N(\cA,h)$ the
number of calls to $L$ that $\cA$ makes before outputting
$(L(x_1),\ldots,L(x_m))$, under the assumption that $L \determined h$.
The worst-case label complexity of $\cA$ is defined to be 
\[
\wCost(\cA) \eqdef \max_{h \in \H} N(\cA,h).
\]
We denote the optimal worst-case
label complexity for the given pool by $\OPT_{\max}$. Formally, we define $\OPT_{\max} = \min_{\cA} \wCost(\cA)$, where the minimum is taken over all possible active learners for the given pool.

Given a probability distribution $P$ over $\cH$, the average-case label complexity of $\cA$ is defined to be
\[
\avgCost(\cA) \eqdef \E_{h \sim P} N(\cA,h).
\] 
The optimal average label complexity for the given pool $X$ and probability distribution $P$ is defined as 
$\OPT_\avg = \min_{\cA} \avgCost(\cA)$. 

For a given active learner, we denote by $V_t \subseteq \H$ the version space of an active learner after $t$ queries. Formally, suppose that the active learning queried instances $i_1,\ldots,i_t$
in the first $t$ iterations. Then
\[
V_t = \{h \in \cH \mid \forall j\in[t], h(x_{i_j}) = L(i_j)\}.
\]
For a given pool example $x \in X$, denote by $V_{t,x}^j$
the version spaces that would result if the algorithm now queried $x$ and received label $j$.
Formally,
\[
V_{t,x}^j = V_t \cap \{h \in \cH \mid h(x) = j\}.
\]

A greedy algorithm (with respect to a probability distribution $P$) is an algorithm $\cA$ that at each iteration $t=1,\ldots,T$,
the pool example $x$ that $\cA$ decides to query is one that splits the version space
as evenly as possible. Formally, at every iteration $t$ $\cA$ queries some example in $\argmin_{x \in X} \max_{j \in \{\pm 1\}} P(V_{t,x}^j)$.
Equivalently, a greedy algorithm is an algorithm $\cA$ that at every iteration $t$ queries an example in 
\[
\argmax_{x \in X}P(V_{t,x}^{-1}) \cdot P(V_{t,x}^{+1}).
\]
To see the equivalence, note that $P(V_{t,x}^{-1}) =P(V_t) -P(V_{t,x}^{+1})$. 
Therefore, 
\[
P(V_{t,x}^{-1}) \cdot P(V_{t,x}^{+1}) = (P(V_t) - P(V_{t,x}^{+1}))P(V_{t,x}^{+1}) = 
(P(V_t)/2)^2 - (P(V_t)/2 - P(V_{t,x}^{+1}))^2.
\]
It follows that the expression is monotonic decreasing in $|P(V_t)/2 - P(V_{t,x}^{+1})|$.

This equivalent formulation motivates the following definition of an approximately greedy algorithm, following \cite{Golovin}. 
\begin{definition}
An algorithm $\cA$ is called \emph{$\alpha$-approximately greedy} with respect to $P$, for $\alpha \geq 1$,
if at each iteration $t=1,\ldots,T$, the pool example $x$ that $\cA$ decides to query satisfies
\[
P(V_{t,x}^1) P(V_{t,x}^{-1}) \geq \frac{1}{\alpha}\max_{\tilde{x} \in X}P(V_{t,\tilde{x}}^1) P(V_{t,\tilde{x}}^{-1}),
\]
and the output of the algorithm is $(h(x_1),\ldots,h(x_m))$ for some $h \in V_T$.
\end{definition}
It is easy to see that by this definition, an algorithm is exactly greedy if it is approximately greedy with $\alpha = 1$.

By \cite{Dasgupta05} we have the following guarantee: For any exactly greedy algorithm $\cA$ with respect to distribution $P$, 
\[
\avgCost(\cA) = O(\log(1/p_{\min})\cdot \OPT_\avg).
\]
\cite{Golovin} show that for an $\alpha$ approximately greedy algorithm, 
\[
\avgCost(\cA)  = O(\alpha \cdot \log(1/p_{\min})\cdot \OPT_\avg).
\]
In addition, they show a similar bound for the worst-case label complexity. Formally,
\begin{equation}\label{eq:golovinbound}
\wCost(\cA) = O(\alpha \cdot \log(1/p_{\min})\cdot \OPT_{\max}).
\end{equation}

\section{Results for Greedy Active Learning} \label{sec:main}

The approximation factor guarantees cited above all inversely depend on $p_{\min}$, the smallest probability
of any hypothesis in the given hypothesis class, according to the given distribution. Thus, if $p_{\min}$ is very small,
the approximation factor is large, regardless of the true target hypothesis. We show that by slightly changing
the policy of an approximately-greedy algorithm, we can achieve a better
approximation factor whenever the true target hypothesis has a larger probability than $p_{\min}$.
This can be done by allowing the algorithm to stop before it reaches a pure version space, that is before it can be certain of the correct labeling of the pool, and requiring that in this case, it would output the labeling which is most likely based on the current version space and the fixed probability distribution $P$. We say that $\cA$ \emph{outputs an approximate majority vote} if whenever $V_T$ is pure enough, the algorithm outputs the majority vote on $V_T$.
Formally, we define this as follows.
\begin{definition} 
An algorithm $\cA$ outputs a $\beta$-approximate majority vote for $\beta \in (\half, 1)$ if whenever there exists a labeling
$Z:X \rightarrow \{\pm1\}$ such that $\prob_{h \sim P}[Z\determined h \mid h \in V_T] \geq \beta$,
$\cA$ outputs $Z$.
\end{definition}
In the following theorem we provide the target-dependent label complexity
bound, which holds for any approximate greedy algorithm that outputs
an approximate majority vote. We give here a sketch of the proof idea, the complete proof can be found in \appref{sec:generalProof}. 
\begin{theorem} \label{thm:general} Let $X=\{x_1, \ldots, x_m\}$. Let $\H$ be a hypothesis class, and let $P$ be a
  distribution over $\H$. Suppose that $\cA$ is $\alpha$-approximately greedy with respect to $P$.
  Further suppose that it outputs a $\beta$-approximate majority vote. If $\cA$ is executed with input $(X,L,T)$ where $L\determined h \in \H$, then for all  
  \[
  T \geq \alpha(2\ln(1/P(h)) + \ln(\frac{\beta}{1-\beta}))\cdot\OPT_{\max},
  \]
  $\cA$ outputs $L(1),\ldots,L(m)$.
 \end{theorem}	
 \begin{proof}[Sketch]
 Fix a pool $X$. 
For any algorithm $\alg$, denote by $V_t(\alg,h)$ the version space induced by the first $n$ labels it queries if the true labeling of the pool is consistent with $h$. 
Denote the average version space reduction of $\alg$ after $t$ queries by 
\[
f_\avg(\alg, t) = 1-\E_{h \sim P}[P(V_t(\alg, h))].
\]
\citep{Golovin} prove that since $\cA$ is $\alpha$-approximately greedy,
for any pool-based algorithm $\alg$, and for every $k,t \in
  \mathbb{N}$, 
\begin{equation}\label{eq:gk}
f_\avg(\cA, t) \geq f_\avg(\alg,k) (1- \exp(-t/\alpha k)).
\end{equation}
Let $\opt$ be an algorithm that achieves $\OPT_{\max}$. We show (see \appref{sec:generalProof}) that for any hypothesis $h \in \H$ and any active learner $\alg$,
\[
f_\avg(\opt,\OPT_{\max}) - f_\avg(\alg, t) \geq P(h)(P(V_t(\alg,h)) - P(h)).
\]
Combining this with \eqref{eq:gk} we conclude that if $\cA$ is $\alpha$-approximately greedy then
\[
\frac{P(h)}{P(V_t(\cA,h))} \geq \frac{P(h)^2}{\exp(-\tfrac{t}{\alpha \OPT_{\max}}) + P(h)^2}.
\]
This means that if $P(h)$ is large enough and we run an approximate greedy algorithm, then after a sufficient number of iterations, most of the remaining version space induces the correct labeling of the sample. 
Specifically, if 
$
t \geq \alpha(2\ln(1/P(h)) + \ln(\frac{\beta}{1-\beta}))\cdot\OPT_{\max},
$
then $P(h)/P(V_t(\cA,h)) \geq \beta$. 
Since $\cA$ outputs a $\beta$-approximate majority labeling from $V_t(\cA,h)$,
$\cA$ returns the correct labeling.
\end{proof}

When $P(h) \gg p_{\min}$, the bound in \thmref{thm:general} is
stronger than the guarantee in \eqref{eq:golovinbound}, obtained by
\citep{Golovin}.  Note, however, that this bound depends on the
probability of the target hypothesis and thus is not known a-priori,
unless additional assumptions are made. The margin assumption, which
we discuss below, is an example for such a plausible
assumption. Moreover, our experimental results indicate that even when
such an apriori bound is not known, using a majority vote is preferable to selecting an arbitrary random hypothesis from an impure version space (see \figref{fig:qbcRandVsMaj} in \secref{sec:experiments}).

Importantly, such an improved approximation factor \emph{cannot be 
obtained} for a general approximate-greedy algorithm, even in a very simple setting. Thus, we can conclude that some algorithmic change is necessary. To show this, consider the setting of \emph{thresholds on the line}. In this setting, the domain of examples is
 $[0,1]$, and the hypothesis class includes all the hypotheses
 defined by a threshold on $[0,1]$. Formally, 
 \[
 \H_\thr = \{ h_c \mid c \in  [0,1], h_c(x) = 1 \Leftrightarrow x \geq c\}.
 \]
 Note that this setting is isomorphic to the case of homogeneous halfspaces with examples on a line in any Euclidean space of two or more dimensions.
   
\begin{theorem}\label{thm:approxline}
Consider pool-based active learning on $\H_\thr$, and assume that $P$ on $\H_\thr$ selects $h_c$ by drawing the value $c$ uniformly from $[0,1]$.
For any $\alpha > 1$ there exists an $\alpha$-approximately greedy
algorithm $\cA$ such that for any $m > 0$ there exists a pool $X
\subseteq [0,1]$ of size $m$, and a threshold $c$ such that $P(h_c) =
1/2$, while the label-complexity of $\cA$ for $L\determined h_c$ is
$\frac{m}{\ceil{\log(m)}}\cdot\OPT_{\max}$.
\end{theorem}
\begin{proof}
For the hypothesis class $\H_\thr$,
the possible version spaces after a partial run of an active learner
are all of the form $[a,b] \subseteq [0,1]$. 

First, it is easy to see that binary search on the pool can identify any hypothesis in $[0,1]$ using $\ceil{\log(m)}$ example, thus $\OPT_{\max} = \ceil{\log(m)}$.
Now, Consider an active learning algorithm that satisfies the following properties:
\begin{itemize}
\item If the current version space is $[a,b]$, it queries the smallest $x$ that would still
make the algorithm $\alpha$-approximately greedy. Formally, it selects 
\[
x = \min\{x \in X \mid (x-a)(b-x) \geq \frac{1}{\alpha}\max_{\tilde{x} \in X\cap [a,b]} (\tilde{x}-a)(b- \tilde{x}) \}.
\]
\item When the budget of queries is exhausted, if the version space is $[a,b]$,
then the algorithm labels the points above $a$ as positive and the rest as negative.
\end{itemize}
It is easy to see that this algorithm is $\alpha$-approximately greedy, 
since in this problem $V_{t,x}^1 \cdot V_{t,x}^{-1} = (x-a)(b-x)$ for all $x \in [a,b] = V_t$. 
Now for a given pool size $m \geq 2$, consider a pool of examples defined as follows. First, let $x_1 = 1$,  $x_2 = 1/2$ and $x_3 = 0$. Second, for each $i \geq 3$, define $x_{i+1}$ recursively as the solution to $(x_{i+1}-x_i)(1-x_{i+1}) = \frac{1}{\alpha}(x_2-x_i)(x_1-x_2)$. Since $\alpha > 1$, it is easy to see by induction that for all $i \geq 3$, $x_{i + 1} \in (x_i, x_2)$. Furthermore,
suppose the true labeling is induced by $h_{3/4}$; Thus the only pool example with a positive label is $x_1$,
and $P(h_{3/4}) = 1/2$. In this case, 
the algorithm we just defined will query all the pool examples $x_4,x_5,\ldots,x_m$ in order, and only then will it query $x_2$ and finally $x_1$. If stopped at any time $t \leq m-1$, it will 
label all the points that it has not queried yet as positive, thus if $t < m-1$ the output will be an erroneous labeling.
Finally, note that the same holds for the pool $x_1,x_2,x_4,\ldots,x_m$ that does not include $x_3$, so the algorithm must query this entire pool to identify the correct labeling.
\end{proof}
Interestingly, this theorem does not hold for $\alpha = 1$, that is for the exact greedy algorithm. This follows from \thmref{thm:line}, which we state and prove in \secref{sec:comparison}.

  So far we have considered a general hypothesis class. We
now discuss the class of halfspaces in $\reals^d$, denoted by $\cW$ above. For simplicity,
we will slightly overload notation and sometimes use $w$ to denote the
halfspace it determines. Every hypothesis in $\cW$ can be described by a vector $w \in \ball^d$, where $\ball^d$ is the Euclidean unit ball, $\ball^d = \{ w \in \reals^d \mid \|w\|\leq 1\}$.
We fix the distribution $P$ to be the one that selects a vector $w$
uniformly from $\ball^d$. Our active learning algorithm for halfspaces, which is called ALuMA, is presented in \secref{sec:alg}. ALuMA receives as
input an extra parameter $\delta \in (0,1)$, which serves as a measure of the
desired confidence level. The following lemma, which we prove in \secref{sec:alg}, shows that ALuMA has the desired
properties described above with high probability.
  
\begin{lemma}\label{lem:aluma}
If ALuMA is executed with confidence $\delta$, then with probability $1-\delta$ over its internal randomization, ALuMA is $4$-approximately greedy and outputs a $2/3$-approximate majority vote.
Furthermore, ALuMA is polynomial in the pool size, the dimension, and $\log(1/\delta)$. 
\end{lemma}

Combining the above lemma with \thmref{thm:general} we immediately
obtain that ALuMA's label complexity is
$O(\log(1/P(h))\cdot\OPT_{\max})$. We can upper-bound $\log(1/P(h))$
using the familiar notion of \emph{margin}: For any hypothesis $h\in \cW$ defined by some $w \in
\ball^d$, let $\gamma(h)$ be the maximal margin of the labeling of
$X$ by $h$, namely $\gamma(h) = \max_{v : \|v\|=1} \min_{i\in [m]}
h(x_i)\inner{v,x_i}/\|x_i\|$. We have the following lemma, which we prove in \appref{app:other}:

\begin{lemma}\label{lem:phbound}
For all $h \in \cW$, $P(h) \geq \left(\frac{\gamma(h)}{2}\right)^d.$
\end{lemma}

From \lemref{lem:phbound} and \lemref{lem:aluma}, we obtain the following corollary,
which provides a guarantee for ALuMA that depends on the margin of the target hypothesis.
\begin{corollary} \label{cor:aluma} Let $X=\{x_1, \ldots, x_m\} \subseteq
  \ball^d$, where $\ball^d$ is the unit Euclidean ball of
  $\reals^d$.  Let $\delta
  \in (0,1)$ be a confidence parameter. Suppose that ALuMA is executed with input $(X,L,T,\delta)$, where $L \determined h \in \cW$ and $T \ge 4(2d\ln(2/\gamma(h)) + \ln(2))\cdot\OPT_{\max}$. Then, with
  probability of at least $1-\delta$ over ALuMA's own randomization,
  it outputs $L(1),\ldots,L(m)$.
 \end{corollary}

Note that ALuMA is allowed to
  use randomization, and it can fail to output the correct label with
  probability $\delta$. In contrast, in the definition of $\OPT_{\max}$ we
  required that the optimal algorithm always succeeds, in effect making it deterministic. One may suggest that
  the approximation factor we achieve for ALuMA in \lemref{lem:aluma}
  is due to this seeming advantage for ALuMA.  We now show that this is not the case---the same approximation factor can be achieved when ALuMA and the optimal algorithm are allowed the same probability of failure.
  Let $m$ be the size of the pool and let $d$ be the dimension of the examples,
  and set $\delta_0 = \frac{1}{2m^d}$. Denote by $N_\delta(\cA,h)$ the
	number of calls to $L$ that $\cA$ makes before outputting
	$(L(x_1),\ldots,L(x_m))$ with probability at least $1-\delta$, for $L \determined h$.
  Define $\OPT_{\delta_0} = \min_A \max_h N_{\delta_0}(A,h)$.
  
  First, note that by setting $\delta = \delta_0$ in ALuMA, we get that $N_\delta(\mathrm{ALuMA}, h) \leq O(\log(1/P(h))\cdot \OPT_{\max})$. Moreover, ALuMA with $\delta = \delta_0$ is polynomial in $m$ and $d$ (since it is polynomial in $\ln(1/\delta)$). Second, by Sauer's lemma there are at most $m^d$ different
  possible labelings for the given pool. Thus by the union bound, there exists a fixed choice
  of the random bits used by an algorithm that achieves $\OPT_{\delta_0}$, that leads to the correct identification of the labeling for \emph{all} possible labelings $L(1),\ldots,L(m)$. It follows that $\OPT_{\delta_0} = \OPT_{\max}$. 
	Therefore the same factor of approximation can be achieved for
        ALuMA with $\delta = \delta_0$, compared to
        $\OPT_{\delta_0}$.

Our result for ALuMA provides a target-dependent approximation factor guarantee, depending on the margin of the target hypothesis. 
We can also consider the minimal possible
margin, $\gamma = \min_{h\in \cW} \gamma(h)$, and deduce from
\corref{cor:aluma}, or from the results of \citep{Golovin}, a uniform
approximation factor of $O(d \log(1/\gamma))$.  How small can $\gamma$ be? The
following result bounds this minimal margin from below under the reasonable assumption that the examples are represented by numbers of a finite accuracy.
 \begin{lemma} \label{lem:grid}
 Let $c > 0$ be such that $1/c$ is an integer and suppose that $X
 \subset \{-1,\allowbreak -1\nobreak +\nobreak c,\allowbreak \ldots,1\nobreak-\nobreak c,1\}^d$. Then, $\min_{h\in \cW} \gamma(h) \ge
 (c/\sqrt{d})^{d+2}$.
 \end{lemma}
 The proof, given in \appref{app:other}, is an adaptation of a classic
 result due to \citep{MurogaToTa61}. We conclude that under this assumption for halfspaces, $p_{\min} =
 \Omega((c/d)^{d^2})$, and deduce an approximation factor of
 $d^2\log(d/c)$ for the worst-case label complexity of ALuMA.
 The exponential dependence of the minimal margin on $d$ here is
 necessary; as shown in \citep{Haastad}, the minimal margin can indeed be exponentially small, even if the points are taken only from $\{\pm 1\}^d$.

 We also derive a lower bound, showing that the dependence of our bounds on $\gamma$ or on $c$
 is necessary. Whether the dependence on $d$ is also necessary is an open question for future work.
\begin{theorem} \label{thm:lower-bound} For any $\gamma \in (0,1/8)$,
   there exists a pool $X \subseteq \ball^2 \cap \{-1,1+c, \ldots,
   1-c,1\}^2$ for $c = \Theta(\gamma)$, and a target hypothesis $h^* \in \mathcal{W}$ for which $\gamma(h^*) = \Omega(\gamma)$, such that there exists an exact greedy algorithm
   that requires $\Omega(\ln(1/\gamma)) = \Omega(\ln(1/c))$ labels in order to output a correct majority vote, while the optimal algorithm requires only $O(\log(\log(1/\gamma)))$ queries.
 \end{theorem}
The proof of \thmref{thm:lower-bound} is provided in \appref{app:other}.
In the next section we describe the ALuMA algorithm in detail.

 \section{The ALuMA algorithm} \label{sec:alg}
 We now describe our algorithm, listed below as \algref{algo:SSA}, and explain why \lemref{lem:aluma} holds. We name the algorithm \emph{Active Learning under a Margin Assumption} or ALuMA.
  Its inputs are the unlabeled sample $X$, the labeling oracle $L$, the maximal allowed number 
 of label queries $T$, and the desired confidence $\delta \in (0,1)$. It returns the labels of all the examples in $X$. 
	
As we discussed earlier, in each iteration, we wish to choose among
the instances in the pool, the instance whose label would
lead to the maximal (expected) reduction in the version space. Denote by $I_t$
the set of indices corresponding to the elements in the pool whose
label was not queried yet ($I_0 = [m]$). Then, in round $t$, we wish to find

% \[
% k = \argmin _{i \in I_t} \max \{P(V_{t,x_i}^1) , P(V_{t,x_i}^{-1})\}.
% \] 

% Equivalently, we look for 

\begin{equation}  \label{eq:exactGreedy}
k = \argmax _{i \in I_t} P(V_{t,x_i}^1) \cdot P(V_{t,x_i}^{-1}). 
\end{equation}

Recall we take $P$ to be uniform over $\mathcal{W}$, the class
of homogenous half-spaces in $\reals^d$. In this case, the probability of a version space is equivalent to its volume, up to constant factors. Therefore, in order to be able to solve \eqref{eq:exactGreedy}, we need to calculate the
 volumes of the sets $V_{t,x}^1$ and $V_{t,x}^{-1}$ for every element
 $x$ in the pool. Both of these sets are convex sets obtained by intersecting the unit ball with
 halfspaces. The problem of calculating the volume of such convex sets
 in $\reals^d$ is \#P-hard if $d$ is not fixed \citep{Brightwell}. In many learning applications $d$ is large, therefore, indeed d should not be taken as fixed.
 Moreover, deterministically approximating the volume is NP-hard in the
 general case \citep{Matousek}. Luckily, it is possible to approximate
 this volume using randomization. 
Specifically, in \cite{Lovasz} a
randomized algorithm with the following guarantees is provided, where $\Vol(K)$ denotes the volume of the set $K$.
\begin{lemma} \label{lm:est}
Let $K \subseteq \reals^d$ be a convex body with an efficient
separation oracle. There exists a randomized algorithm, such that
given $\epsilon, \delta > 0$, with probability at least $1-\delta$ the
algorithm returns a non-negative number $\Gamma$ such that 
$
(1-\epsilon)\Gamma < \Vol(K) < (1+\epsilon) \Gamma.
$
The running time of the algorithm is polynomial in $d, 1/\epsilon,\ln(1/ \delta)$.
\end{lemma}

We denote an execution of this algorithm on a convex
 body $K$ by $\Gamma \leftarrow \VolEst(K,\epsilon,\delta)$. The algorithm is polynomial in $d, 1/\epsilon,\ln(1/ \delta)$.
 ALuMA uses this algorithm to estimate $P(V^1_{t,x})$ and
 $P(V^{-1}_{t,x})$ with sufficient accuracy. We denote these
 approximations by $\hat{v}_{x,1}$ and $\hat{v}_{x,-1}$ respectively. Using the constants in ALuMA, we can show the following.
\begin{lemma}  \label{lem:greedySSA}
With probability at least $1-\delta/2$, \algref{algo:SSA} is $4$-approximately greedy.
\end{lemma}

\begin{proof}
Fix some $t \in [T]$. Let $k \in I_t$ be the index chosen by ALuMA. Let $k^*$ be the
index corresponding to the value of \eqref{eq:exactGreedy}. Since ALuMA performs at most $2m$ approximations in each round, we obtain
by \lemref{lm:est} and the union bound that with probability at
least $1-\frac{\delta}{2T}$, for each $i \in I_t$ and each $j \in \{-1,1\}$, 
\[
\hat{v}_{x_i,j}  \in \left(\frac{2}{3} \Vol(V_{t,x_i} ^j), \frac{4}{3}
    \Vol(V_{t,x_i} ^j) \right).
\]
In addition, $\hat{v}_{x_k,1}\cdot\hat{v}_{x_k,-1}\geq \hat{v}_{x_{k^*},1}\cdot\hat{v}_{x_{k^*},-1}$.
Hence, with probability at
least $1-\frac{\delta}{2T}$, 

\[
\frac{16}{9} \Vol(V_{t,x_k}^{-1}) \cdot \Vol(V_{t,x_k}^1) \geq \frac{4}{9} \Vol(V_{t,x_{k^*}}^{-1}) \cdot \Vol(V_{t,x_{k^*}}^1).
\]

Applying the union bound over $T$ iteration completes our proof.
\end{proof}

\begin{algorithm}[t] 
 \caption{The \textbf{ALuMA} algorithm} \label{algo:SSA}
 \begin{algorithmic}[1]
 \STATE {\bf Input:} $X=\{x_1, \ldots, x_m\}$, $L:[m] \rightarrow \{-1,1\}$, $T$, $\delta$
 \STATE  $I_1 \leftarrow [m]$, $V_1 \leftarrow \ball^d$
 \FOR {$t=1$ to $T$}
 \STATE $\forall i \in I_t, j \in \{\pm 1\}$, do $\hat{v}_{x_i,j} \leftarrow \VolEst(V_{t,x_i}^{j},\frac{1}{3},\frac{\delta}{4m T})$\label{step:volest}
 \STATE Select $i_t \in \argmax_{i \in I_t} (\hat{v}_{x_i,1} \cdot \hat{v}_{x_i,-1})$\label{step:select}
 \STATE $I_{t+1} \leftarrow  I_t \setminus \{i_t\}$
 \STATE Request $y=L(i_t)$
 \STATE $V_{t+1} \leftarrow V_t \cap \{w : y\inner{w,x_{i_t}} > 0\}$
 \ENDFOR
 \STATE $M \leftarrow \lceil 72\ln(2/\delta) \rceil$.
 \STATE Draw $w_1,\ldots,w_M$ $\frac{1}{12}$-uniformly from $V_{T+1}$.\label{step:draw}
 \STATE For each $x_i$ return the label $y_i = \sgn\left(\sum_{j=1}^M \sgn(\inner{w_j,x_i})\right)$.
 \end{algorithmic}
 \end{algorithm} 
	
	After $T$ iterations, ALuMA needs to output the majority vote of a version space that has a high enough purity level.
To output an approximate majority vote from the final version space
$V$, we would like to uniformly draw several hypotheses from $V$ and
label $X$ according to a majority vote over these hypotheses. 
The task of uniformly drawing hyphteses from $V$ can be approximated
using the hit-and-run algorithm \citep{lovasz1999hit}. The hit-and-run algorithm efficiently draws a random sample from a convex body $K$ according to a distribution which is close in total variation distance to the uniform distribution over $K$. Formally,
The following definition parametrizes the closeness of a distribution to the uniform distribution:
\begin{definition} \label{def:lambdaUnfiormly}
Let $K \subseteq \reals^d$ be a convex body with an efficient separation oracle, and let $\tau$ be a distribution over $K$. $\tau$ is  \emph{$\lambda$-uniform} if $
\sup_{A} |\tau(A)-P(A)/P(K)| \leq \lambda,
$
where the supremum is over all measurable subsets of $K$.
\end{definition}

The hit-and-run algorithm draws a sample from a $\lambda$-uniform
distribution in time $\tilde{O}(d^3/\lambda^2)$. The next lemma shows
that using the hit-and-run as suggested above indeed produces a
majority vote classification.

\begin{lemma}  \label{lem:majSSA}
ALuMA outputs a $2/3$-approximate majority vote with probability at least
$1-\delta/2$.
\end{lemma}

\begin{proof}
Assume that there exists a labeling
$Z:X \rightarrow \{\pm1\}$ such that 
$
\prob_{h \sim P}[Z\determined h \mid h \in V_{T+1}] \geq 2/3.
$
 In step \ref{step:draw} of ALuMA, $M \geq 72\ln(2/\delta)$ hypotheses are drawn
$\frac1{12}$-uniformly at random from $V_t$. Therefore each hypothesis
$h_i \in V_{T+1}$ is consistent with $Z$ with probability at least $\frac{7}{12}$. By Hoeffding's
inequality, 
\[
\prob \left[\frac{1}{M}\sum_{i=1}^{M} I[h_i \in V(h|_X)] \leq \half \right] \leq
\exp(-M/72)  = \frac{\delta}{2}.
\]
Therefore, with probability at least $1-\delta/2$, ALuMA outputs a
$2/3$-approximate majority vote.

\end{proof}

We can now prove \lemref{lem:aluma}.
\begin{proof} \textbf{(Of \lemref{lem:aluma})}
\lemref{lem:greedySSA} and \lemref{lem:majSSA} above prove the first two parts of the lemma. We only have left to analyze the time complexity of ALuMA. In each iteration, the
cost of ALuMA is dominated by the cost of performing at most $2m$ volume approximation, each of which costs $O(d^5
\ln (1/\delta))$. As we discussed above, implementing the majority vote
costs polynomial time in $d$ and $\ln(1/\delta)$. Overall, the runtime
of ALuMA is polynomial in $m$ (which upper bounds $T$), $d$ and
$\log(1/\delta)$. 
\end{proof}

\subsection{A Simpler Implementation of ALuMA}\label{sec:simpler}

The ALuMA algorithm described in \algref{algo:SSA} uses $O(Tm)$ volume estimations as a black-box procedure, where $T$ is the budget of labels and $m$ is the pool size. The complexity of each application of the volume estimation procedure is $\tilde{O}(d^5)$ where $d$ is the dimension. Thus the overall complexity of the algorithm is $\tilde{O}(Tm d^5 )$. This complexity can be somewhat improved under some ``luckiness'' conditions.

The volume estimation procedure uses $\lambda$-uniform sampling based on hit-and-run as its core procedure. Instead, we can use hit-and-run directly
as follows: At each iteration of ALuMA, instead of step \ref{step:volest}, perform the following procedure:

\begin{algorithm}[th] 
\caption{Estimation Procedure}\label{alg:estimation}
\begin{algorithmic}[1]
\STATE Input: $\lambda \in (0,\frac{1}{24}),V_t,I_t$
\STATE $k \leftarrow \frac{\ln(2Nm/\delta)}{2\lambda^2}$ 
\STATE Sample $h_1,\ldots,h_k \in V_t$ $\lambda$-uniformly. 
\STATE $\forall i \in I_t, j \in \{-1,+1\}$, $\hat{v}_{x_i,j} \leftarrow \frac{1}{k}|\{i \mid h_i(x_i) = j\}|$.
\end{algorithmic}
\end{algorithm}

The complexity of ALuMA when using this procedure is
$\tilde{O}(T(d^3/\lambda^4 + m/\lambda^2))$, which is better than the
complexity of the full \algref{algo:SSA} for a constant $\lambda$. An
additional practical benefit of this alternative estimation procedure
is that when implementing, it is easy to limit the actual computation
time used in the implementation by running the procedure with a
smaller number $k$ and a smaller number of hit-and-run mixing
iterations.\footnote{\cite{KQBC} report that the actual mixing time of
  hit-and-run is much faster than the one guaranteed by the
  theoretical bounds, and we have observed a similar phenomenon in our
  experiments.} This provides a natural trade-off between computation
time and labeling costs.

The following theorem shows that under mild conditions, using the estimation procedure listed
in \algref{alg:estimation} also results in an approximately greedy
algorithm, as does the original implementation of ALuMA.

\begin{theorem}   \label{thm:simplerAnalysis}
If for each iteration $t$ of the algorithm, the greedy choice $x^*$ 
satisfies 
\[
\forall j \in \{-1,+1\},\quad \prob[h(x^*) = j \mid h \in V_t] \geq 4\sqrt{\lambda}
\]
then ALuMA with the estimation procedure is a $2$-approximate greedy algorithm.
Moreover, it is possible to efficiently verify that this condition holds while running the algorithm.
\end{theorem}

\begin{proof}
Fix the iteration $t$, and denote $p_{x,1} = P(V^1_{t,x})/P(V_t)$ and 
$p_{x,-1} = P(V^1_{t,x})/P(V_t)$. Note that $p_{x,1} + p_{x,-1} = 1$.
Since $h_1,\ldots,h_k$ are sampled $\lambda$-uniformly from the version space, 
we have 
\begin{equation}\label{eq:lambda1}
\forall i \in [k], |\prob[h_i \in V^j_{t,x}] - p_{x,j}| \leq \lambda.
\end{equation}
In addition, by Hoeffding's inequality and a union bound over the examples in the pool and the iterations of the algorithm, 
\begin{equation}\label{eq:lambda2}
\prob[\exists x, |\hat{v}_{x_i,j} - \prob[h_i \in V^j_{t,x}]| \geq \lambda] \leq 2m\exp(-2k\lambda^2).
\end{equation}
From \algref{alg:estimation} we have  $k = \frac{\ln(2m/\delta)}{2\lambda^2}$. Combining this with \eqref{eq:lambda1} and \eqref{eq:lambda2} we get
that 	
\[
\prob[\exists x, |\hat{v}_{x_i,j} - p_{x_i,j}]| \geq 2\lambda] \leq \delta.
\]

The greedy choice for this iteration is
\[
x^* \in \argmax_{x \in X} \Delta(h|_X,x) = \argmax_{x \in X}(p_{x,1} p_{x,-1}).
\]

By the assumption in the theorem,
$p_{x^*,j} \geq 4\sqrt{\lambda}$ for $j \in \{-1,+1\}$.
Since $\lambda \in (0,\frac{1}{64})$, we have $\lambda \leq \sqrt{\lambda}/8$.
Therefore
$p_{x^*,j} -2\lambda \geq 4\sqrt{\lambda} - \sqrt{\lambda}/4 \geq \sqrt{10\lambda}$. 
Therefore
\begin{equation}\label{eq:cond2}
\hat{v}_{x^*,1}\hat{v}_{x^*,-1} \geq  (p_{x^*,1} - 2\lambda)(p_{x^*,-1} - 2\lambda)
\geq 10\lambda.
\end{equation}

Let $\tilde{x} = \argmax (\hat{v}_{x,-1}\hat{v}_{x,+1})$ be the query selected by ALuMA using \algref{alg:estimation}. 
Then
\begin{align*}
\hat{v}_{x^*,-1}\hat{v}_{x^*,+1} \leq
\hat{v}_{\tilde{x},-1}\hat{v}_{\tilde{x},+1}\leq (p_{\tilde{x},1}+2\lambda)(p_{\tilde{x},-1}+2\lambda) \leq p_{\tilde{x},1}p_{\tilde{x},-1} + 4\lambda.
\end{align*}
Where in the last inequality we used the facts that $p_{\tilde{x},1}+p_{\tilde{x},-1}=1$ and $4\lambda^2 \leq 2\lambda$.
On the other hand, by \eqref{eq:cond2}
\begin{align*}
\hat{v}_{x^*,-1}\hat{v}_{x^*,+1} \geq 5\lambda + \half\hat{v}_{x^*,-1}\hat{v}_{x^*,+1} \geq 
5\lambda + \half(p_{x^*,-1}-2\lambda)(p_{x^*,-1}-2\lambda)
\geq 4\lambda + \half p_{x^*,-1}p_{x^*,-1}.
\end{align*}

Combining the two inequalities for $\hat{v}_{x^*,-1}\hat{v}_{x^*,+1}$ it follows that $p_{\tilde{x},1}p_{\tilde{x},-1} \geq \half p_{x^*,-1}p_{x^*,-1}$,
thus this is  a $2$-approximately greedy algorithm.

To verify that the assumption holds at each iteration of the algorithm,
note that for all $x = x_i$ such that $i\in I_t$
\[
p_{x,-1}p_{x,+1} \geq (\hat{v}_{x,-1} - 2\lambda)(\hat{v}_{x,+1} - 2\lambda) \geq \hat{v}_{x,-1}\hat{v}_{x,+1} - 2\lambda.
\]
therefore it suffices to check that for all $x = x_i$ such that $i\in I_t$
$\hat{v}_{x,-1}\hat{v}_{x,+1} \geq 4\sqrt{\lambda} + 2\lambda.$
\end{proof}

The condition added in this theorem is that the best example
in each iteration should induce a fairly balanced partition of the current version
space. In our experiments we noticed that this is generally the case in
practice. Moreover, the theorem shows that it is possible to verify that the
condition holds while running the algorithm. Thus, the estimation procedure can
easily be augmented with an additional verification step at the beginning of
each iteration. On iterations that fail the verification, the algorithm will use the original 
black-box volume estimation procedure. We have used this simpler implementation in our experiments, which are reported below.

\subsection{Handling Non-Separable Data and Kernel Representations}
\label{sec:transformation}
If the data pool $X$ is not separable, but a small upper bound on the
total hinge-loss of the best separator can be assumed, then ALuMA can
be applied after a preprocessing step, which we describe in detail below.  This preprocessing step maps the points in $X$ to a set of points in a higher dimension, which are separable using the original labels of $X$. The dimensionality depends on the margin and on the bound on the total hinge-loss of the original representation. The preprocessing step  also supports kernel representations, so that the original $X$ can be represented by a kernel matrix as well. Applying ALuMA after this preprocessing steps results in an approximately optimal label complexity, however $\OPT_{\max}$ here is measured with respect to the new representation.

While some of the transformations we employ in the preprocessing step have been discussed before in other contexts \citep[see e.g.][]{Blum}, we describe and analyze the full procedure here for completeness. 
The preprocessing step is composed of two simple transformations. In the
first transformation each example $x_i\in X$ is mapped to an example in
dimension $d+m$, defined by $x_i' = (a x_i; \sqrt{1-a^2}\cdot e_i)$,
where $e_i$ is the $i$'th vector of the natural basis of $\reals^m$
and $a>0$ is a scalar that will be defined below. Thus the first $d$ coordinates of $x'_i$ hold the original
vector times $a$, the rest of the coordinates are zero,except for $x'_i[d+i] = \sqrt{1-a^2}$. This mapping
guarantees that the set $X' = (x'_1,\ldots,x'_m)$ is separable with
the same labels as those of $X$, and with a margin that depends on the
cumulative squared-hinge-loss of the data.

In the second transformation, a Johnson-Lindenstrauss random projection \citep{JohnsonLind,Bourgain}
is applied to
$X'$, thus producing a new set of points $\bar{X} =
(\bar{x}_1,\ldots,\bar{x}_m)$ in a different dimension $\reals^k$, where $k$
depends on the original margin and on the amount of margin error. With high
probability, the new set of points will be separable with a margin that also
depends on the original margin and on the amount of margin error.
If the input data is provided not as vectors in $\reals^d$ but via a kernel matrix,
then a simple decomposition is performed before the preprocessing begins.

% If the data $X = \{x_1,\ldots,x_m\}$ is not guaranteed
% to be separable, or it is represented only using a kernel matrix, then
% ALuMA still can be used, after a preprocessing step.  This preprocessing
% step maps the points in $X$ to a set of points in a lower dimension, which are separable using the original labels of $X$. 
 
The full preprocessing procedure is listed below as \algref{algo:preprocess}. The first input to the algorithm is the data for preprocessing, given as $X \subseteq \reals^d$ or as a kernel matrix $K \in \reals^{m\times m}$. The other inputs are $\gamma$ -- a margin parameter, $H$ -- an upper bound on the margin error relative to $\gamma$, and $\delta$, which is the required confidence.
\begin{algorithm}[ht] 
\caption{Preprocessing} \label{algo:preprocess}
\begin{algorithmic}[1]
\STATE {\bf Input:} $X=\{x_1, \ldots, x_m\} \in \reals^d$ or $K \in \reals^{m\times m}$, $\gamma$, $H$, $\delta$
\IF{input data is a kernel matrix $K$}
\STATE Find $U \in \reals^{m \times m}$ such that $K = U U^T$ \label{step:decompose}%there are many ways to do that efficiently. for instance: Cholesky decomposition
\STATE $\forall i \in [m], x_i \leftarrow$ row $i$ of $U$
\STATE $d \leftarrow m$
\ENDIF
\STATE $a \leftarrow \sqrt{\frac{1}{1+\sqrt{H}}}$
\STATE $\forall i \in [m], x'_i \leftarrow (a x_i; \sqrt{1-a^2}\cdot e_i)$\label{step:setxprime}
\STATE $k \leftarrow O\left(\frac{(H+1)\ln(m/\delta)}{\gamma^2}\right)$
\STATE $M \leftarrow $ a random $\{\pm 1\}$ matrix of dimension $k\times(d+m)$
\FOR{$i \in [m]$}
\STATE $\bar{x}_i \leftarrow M x_i'$\label{step:jlproj}
\ENDFOR
\STATE Return $(\bar{x}_1,\ldots, \bar{x}_m).$
\end{algorithmic}
\end{algorithm}

After the preprocessing step, $\bar{X}$ is used as input to ALuMA, which then returns a set of labels for the examples in $\bar{X}$. These are also the labels of the examples in the original $X$. To retrieve a halfspace for $X$ with the least 
margin error, any passive learning algorithm can be applied to the resulting labeled sample. The full active learning procedure is described in \algref{algo:full}. 

Note that if ALuMA returns 
the correct labels for the sample, the usual generalization bounds for
passive supervised learning can be used to bound the true error of the
returned separator $w$. In particular, we can apply the support vector
machine algorithm (SVM) and rely on generalization bounds for SVM. 
\begin{algorithm}[ht] 
\caption{Active Learning} \label{algo:full}
\begin{algorithmic}[1]
\STATE {\bf Input:} $X=\{x_1, \ldots, x_m\}$ or $K \in \reals^{m\times m}$, $L:[m] \rightarrow \{-1,1\}$, $N$, $\gamma$, $H$, $\delta$
\IF{input has $X$}
\STATE Get $\bar{X}$ by running \algref{algo:preprocess} with input $X, \gamma, H, \delta/2$.
\ELSE  
\STATE Get $\bar{X}$ by running \algref{algo:preprocess} with input $K, \gamma, H, \delta/2$.
\ENDIF
\STATE Get $(y_1,\ldots,y_m)$ by running ALuMA with input $\bar{X}$, $L$, $N$, $\delta/2$.
\STATE Get $w \in \reals^d$ by running SVM on the labeled sample $\{(x_1,y_1),\ldots,(x_m,y_m)\}$.
\STATE Return $w$.
\end{algorithmic}
\end{algorithm}

The result of these transformations are summarized in the following
theorem.

\begin{theorem}\label{thm:transform}
Let $X=\{x_1, \ldots, x_m\} \subseteq B$, where $B$ is the unit ball in some
Hilbert space. Let $H \geq 0$ and $\gamma > 0$, and assume there exists a $w^* \in B$
such that
\[
H \geq \sum_{i=1}^m \max(0,\gamma- L(i) \inner{w^*, x_i})^2.
\]
Let $\delta \in (0,1)$ be a confidence parameter.  There exists an algorithm
that receives $X$ as vectors in $\reals^d$ or as a kernel matrix $K \in
\reals^{m \times m}$, and input parameters $\gamma$ and $H$, and outputs a set
$\bar{X} = \{\bar{x}_1,\ldots,\bar{x}_m\} \subseteq \reals^k$, such that
\begin{enumerate}
\item $k = O\left(\frac{(H+1)\ln(m/\delta)}{\gamma^2}\right)$,\label{item:dim}
\item \label{item:separable} With probability $1-\delta$, $\bar{X} \subseteq \ball^k$ and
  $(\bar{X},L)$ is separable with a margin $\frac{\gamma}{2+2\sqrt{H}}$.
\item \label{item:poly2} The run-time of the algorithm is polynomial in $d,m,1/\gamma,\ln(1/\delta)$ if $x_i$ are represented as vectors in $d$, and is polynomial in $m,1/\gamma,\ln(1/\delta)$ if $x_i$ are represented by a kernel matrix.
\end{enumerate}
\end{theorem}
The proof of \thmref{thm:transform} can be found in \appref{app:transformation}. In \secref{sec:experiments} we demonstrate that in practice, this procedure provides good label complexity results on real data sets. Investigating the relationship between $\OPT_{\max}$ in the new representation and $\OPT_{\max}$ in the original representation is an important question for future work.

 \section{Other Approaches: A Theoretical and Empirical Comparison} \label{sec:comparison} We now compare the effectiveness
 of the approach implemented by ALuMA to other active learning
 strategies. ALuMA can be characterized by two properties: (1) its
 ``objective'' is to reduce the volume of the version space and (2) at
 each iteration, it aggressively selects an example from the pool so
 as to (approximately) minimize its objective as much as possible (in
 a greedy sense). We discuss the implications of these properties by
 comparing to other strategies. Property (1) is contrasted with
 strategies that focus on increasing the number of examples whose
 label is known. Property (2) is contrasted with strategies which are
 ``mellow'', in that their criterion for querying examples is softer.

Much research has been devoted to the challenge of obtaining a substantial
 guaranteed improvement of label complexity over regular ``passive'' learning for halfspaces in
 $\reals^d$. Examples (for the realizable case) include the Query By Committee
 (QBC) algorithm \citep{Seung, Tishby}, the CAL algorithm \citep{CAL}, and the Active Perceptron
 \citep{Kalai}.  These
 algorithms are not ``pool-based'' but rather use ``selective-sampling'': they
 sample one example at each iteration, and immediately decide whether to ask
 for its label. 
 Out of these algorithms, CAL is the most mellow, since it queries any example
 whose label is yet undetermined by the version space. Its ``objective'' can be
 described as reducing the number of examples which are labeled incorrectly,
 since it has been shown to do so in many cases \citep{Hanneke07a,Hanneke11,Friedman}. QBC and the active perceptron are less
 mellow. Their ``objective'' is similar to that of ALuMA since they decide on examples to query based on geometric considerations. 

In \secref{sec:theoComp} we discuss the theoretical advantages and
disadvantages of different strategies, by considering some interesting
cases from a theoretical perspective.  In \secref{sec:experiments} we report an empirical comparison of several algorithms and discuss our conclusions.

\subsection{Theoretical Comparison}   \label{sec:theoComp}

% Pool based active learning for PAC
The label complexity of the algorithms mentioned above is usually analyzed in the PAC setting,
thus we translate our guarantees into the PAC setting as well for the sake of
comparison.
We define the $(\epsilon,m,D)$-label complexity of an active learning algorithm to be the number of \emph{label queries} that are required in order to
 guarantee that given a sample of $m$ unlabeled examples drawn from $D$, the error of the learned classifier will be at most $\epsilon$ (with probability of at least $1-\delta$ over the choice of
 sample). A 
a pool-based active learner can be used to learn
 a classifier in the PAC model by first
 sampling a pool of $m$ unlabeled examples from $D$, then applying
 the pool-based active learner to this pool, and finally running a
 standard passive learner on the labeled pool to obtain a
 classifier. For the class of halfspaces, if we sample
 an unlabeled pool of $m = \tilde{\Omega}(d/\epsilon)$ examples, then the learned classifier
 will have an error of at most $\epsilon$ (with high probability over
 the choice of the pool).

To demonstrate the effect of the first property discussed above, consider again
the simple case of thresholds on the line defined in
\secref{sec:main}. Compare two greedy pool-based active learners for $\H_\thr$ :  The first follows a binary search procedure, greedily selecting the example that increases the number
of known labels the most. Such an algorithm requires
$\ceil{\log(m)}$ queries to identify the correct labeling of the pool.
The second algorithm queries the example that 
splits the version space as evenly as possible.
\thmref{thm:general} implies a label complexity of
$O(\log(m)\log(1/\gamma(h)))$ for such an algorithm, since
$\OPT_{\max} = \ceil{\log(m)}$. However, a better result holds for this
simple case: 
\begin{theorem}\label{thm:line}
In the problem of thresholds on the line, for any pool with labeling $L$, the exact greedy algorithm requires at most $O(\log(1/\gamma(h)))$ labels. This is also the label complexity of any approximate greedy algorithm that outputs a majority vote.
\end{theorem}

\begin{proof}
First, assume that the algorithm is exactly greedy. A version space for $\H_\thr$ is described by a segment in $[a,b] \subseteq [0,1]$, and a query at point $\alpha$ results in a new version space, $[a,\alpha]$ or $[\alpha,b]$, depending on the label.
We now show that for every version space $[a,b]$, at most two greedy queries
suffice to either reduce the size of the version space by a factor of at least
$2/3$, or to determine the labels of all the points in the pool. 

Assume for simplicity that the version space is $[0,1]$, and denote the pool of examples in the version space by $X$.
Assume w.l.o.g. that the greedy algorithm now queries $\alpha \leq \half$. 
If $\alpha > 1/3$, then any answer to the query will reduce the version space size to less than $2/3$. Thus assume that $\alpha \leq 1/3$. If the query answer results in the version space $[0,\alpha)$ then we are done since this version space is smaller than $2/3$. We are left with the case that the version space after querying $\alpha$ is $[\alpha,1]$. Since the algorithm is greedy, it follows that for $\beta = \min\{x \in X \mid x \geq \alpha\}$, we have $\beta \geq 1-\alpha$: this is because if there was a point $\beta \in (\alpha, 1-\alpha)$, it would cut the version space more evenly than $\alpha$, in contradiction to the greedy choice of $\alpha$. Note further that $(\alpha, 1-\alpha)$ is larger than $[1-\alpha,1]$ since $\alpha \leq 1/3$. Therefore, the most balanced choice for the greedy algorithm is $\beta$. If the query answer for $\beta$ cuts the version space to $(\beta, 1]$ then we are done, since $1-\beta \leq \alpha \leq 1/3$. Otherwise, the query answer leaves us with the version space $(\alpha,\beta)$. This version space includes no more pool points, by the definition of $\beta$. Thus in this case the algorithm has determined the labels of all points.

It follows that if the algorithm runs at least $t$ iterations, then the size of
the version space after $t$ iterations is at most $(2/3)^{t/2}$. If the true
labeling has a margin of $\gamma$, we conclude that $(2/3)^{t/2} \geq \gamma$,
thus $t \leq O(\log(1/\gamma))$. 

A similar argument can be carried for ALuMA, using a smaller bound on $\alpha$ and more iterations due to the approximation, and noting that if the correct answer is in $(\alpha,1-\alpha)$ then a majority vote over thresholds drawn randomly from the version space will label the examples correctly.
\end{proof}

Comparing the $\ceil{\log(m)}$ guarantee of the first algorithm to the
$\log(1/\gamma(h))$ guarantee of the second, we reach the (unsurprising)
conclusion, that the first algorithm is preferable when the true labeling has a
small margin, while the second is preferable when the true labeling has a large
margin. This simple example accentuates the implications of selecting the
volume of the version space as an objective. A similar implication can be derived by considering the PAC setting, replacing the binary-search algorithm with CAL, and letting $m = \tilde{\Theta}(1/\epsilon)$. On the single-dimensional line, CAL achieves a label-complexity of $O(\log(1/\epsilon)) = O(\log(m))$, similarly to the binary search strategy we described. Thus when $\epsilon$ is large compared to $\gamma(h)$, CAL is better than being greedy on the volume, and the opposite holds when the condition is reversed. QBC will behave similarly to ALuMA in this setting.

To demonstrate the effect of the second property described above---being aggressive versus being mellow, we consider the following example, adapted slightly from \citep{Dasgupta06}. 
\begin{ex}\label{ex:twocircles}
 Consider two circles parallel to the $(x,y)$ plane in $\reals^3$,
one at the origin and one slightly above it.
For a given $\epsilon$, fix $2/\epsilon$ points that are evenly distributed
on the top circle, and $2/\epsilon$ points at the same angles on the bottom circle (see left illustration below). The distribution $D_\epsilon$ is an uneven mix of a uniform distribution over the points on the top circle and one over the points of the bottom circle: The top circle is given a much higher probability. All homogeneous separators label half of the bottom circle positively, but an unknown part of the top circle (see right illustration). The bottom points can be very helpful in finding the correct separator fast, but their probability is low.

%\begin{wrapfigure}{r}{0.5\textwidth}
\begin{center}
\begin{tikzpicture}[scale = 0.6]    
\draw[<-] (0,0,2) -- (0,0,0);
\draw[<-] (0,2,0) -- (0,0,0);
\draw[<-] (2.5,0,0) -- (0,0,0);

\draw[*-*] (-2,0) arc (180:240:2cm and 0.5cm);
\draw[*-*] (-2,0) arc (180:300:2cm and 0.5cm);
\draw[*-*] (-2,0) arc (180:360:2cm and 0.5cm);

\draw[-*] (-2,0) arc (180:120:2cm and 0.5cm);
\draw[-*] (-2,0) arc (180:60:2cm and 0.5cm);
\draw (-2,0) arc (180:0:2cm and 0.5cm);

\draw[*-*] (-1.4,1.2) arc (180:240:1.4cm and 0.3cm);
\draw[*-*] (-1.4,1.2) arc (180:300:1.4cm and 0.3cm);
\draw[*-*] (-1.4,1.2) arc (180:360:1.4cm and 0.3cm);

\draw[-*] (-1.4,1.2) arc (180:120:1.4cm and 0.3cm);
\draw[-*] (-1.4,1.2) arc (180:60:1.4cm and 0.3cm);
\draw (-1.4,1.2) arc (180:0:1.4cm and 0.3cm);
\end{tikzpicture} 
\hspace{4em}  
\begin{tikzpicture} [scale = 0.6]       
\draw[opacity = 0] (0,0,2) -- (0,0,0);
\draw[opacity = 0] (0,2,0) -- (0,0,0);
\draw[opacity = 0] (2.5,0,0) -- (0,0,0);

\draw[*-*] (-2,0) arc (180:240:2cm and 0.5cm);
\draw[*-*] (-2,0) arc (180:300:2cm and 0.5cm);
\draw[*-*] (-2,0) arc (180:360:2cm and 0.5cm);

\draw[-*] (-2,0) arc (180:120:2cm and 0.5cm);
\draw[-*] (-2,0) arc (180:60:2cm and 0.5cm);
\draw (-2,0) arc (180:0:2cm and 0.5cm);

\draw[*-*] (-1.4,1.2) arc (180:240:1.4cm and 0.3cm);
\draw[*-*] (-1.4,1.2) arc (180:300:1.4cm and 0.3cm);
\draw[*-*] (-1.4,1.2) arc (180:360:1.4cm and 0.3cm);

\draw[-*] (-1.4,1.2) arc (180:120:1.4cm and 0.3cm);
\draw[-*] (-1.4,1.2) arc (180:60:1.4cm and 0.3cm);
\draw (-1.4,1.2) arc (180:0:1.4cm and 0.3cm);
%\node at ($(2.3,0)$) {p'};
%\node at ($(-2.3,0)$) {n'};
\node[scale=1.5]  at ($(1.2,0)$) {$+$};
\node[scale=1.5]  at ($(-1.2,0)$) {$-$};

%\node at ($(1.7,2.2)$) {p};
%\node at ($(-1.7,2.2)$) {n};
\node[scale=1] at ($(1,1.2)$) {$+$};
\node[scale=1]  at ($(-0.1,1.2)$) {$-$};

\draw[dashed] (0,-0.5) -- (0,0.5);
\draw[dashed] (0.8,0.95) -- (0.8,1.45);
\end{tikzpicture}
\end{center}
%\end{wrapfigure}
 \end{ex}

Dasgupta has demonstrated via this example that active learning can gain in label complexity from
having significantly more unlabeled data. The following theorem shows
that the aggressive strategy employed by ALuMA indeed achieves an exponential improvement when there are more unlabeled samples. In many applications, unlabeled examples are virtually free to sample, thus it can be worthwhile to allow the active learner to sample more examples than the 
 passive sample complexity and use an aggressive strategy.\footnote{In the limit of an infinite number of unlabeled examples, if the distribution has a non-zero support on the entire domain, the pool-based setting becomes identical to the setting of membership queries \citep{Angluin}. In contrast, we are interested in finite samples.} 
In contrast, the mellow strategy of CAL does not significantly improve over passive learning in this case. We note that these results hold for any selective-sampling method that guarantees an error rate similar to passive ERM given the same sample size.  
This falls in line with the observation of \citep{Balcan07}, that in some cases a more aggressive approach is preferable.

 \begin{theorem}\label{thm:twocircles}
 For all small enough $\epsilon \in (0,1)$ there is a distribution
 $D_\epsilon$ of points in $\reals^3$, such that
\begin{enumerate}
 \item For $m = O(1/\epsilon)$, the $(\epsilon,m,D_\epsilon)$-label complexity of \emph{any active learner} is $\Omega(1/\epsilon)$.
 \item For $m = \Omega(\log ^2 (1/\epsilon)/\epsilon^2)$, the $(\epsilon,m,D_\epsilon)$-label complexity of ALuMA is $O(\log^2(1/\epsilon))$.
 %\item The label complexity of CAL for a sample of \emph{any size} is $\Omega(1/\epsilon)$.
 \item For {\em any value} of $m$, the $(\epsilon,m,D_\epsilon)$-label complexity of CAL is $\Omega(1/\epsilon)$.
\end{enumerate}
\end{theorem}

\begin{proof} 
Assume that $1/(2\epsilon)$ is an odd integer and $\epsilon<1/8$. Let $D_a$ 
be the uniform distribution over points on the top circle, defined by 
\[
S_a = \{a_n \eqdef (\frac{1}{\sqrt{2}}\cos 2 \pi \epsilon n , \frac{1}{\sqrt{2}}\sin 2 \pi \epsilon n,\frac{1}{\sqrt{2}}): ~n \in
\{0,1,\ldots, 1/\epsilon-1\} \}~.
\]
Let $D_b$ be the uniform distribution over points on the bottom circle, defined by 
\[
S_b = \{b_n \eqdef (\cos 2 \pi \epsilon n , \sin 2 \pi \epsilon n,0): ~n \in
\{0,1,\ldots, 1/\epsilon-1\} \}~.
\]
Let $D_{\epsilon/2}$ be the distribution $(1-\tau) D_a + \tau D_b$, where $\tau = \frac{\epsilon}{4 \log  (4/\epsilon)}$. Note that in order to label $D_{\epsilon/2}$ correctly with error
no more than $\epsilon/2$, all the labels of points in $S_a$ need to be determined.
We prove each of the theorem statements in order. We consider the label complexity with high probability over the choice of unlabeled sample, where high probability is $1-\delta$ for some fixed $\delta \in (0,1/2)$.
\paragraph{Part 1.}
If the unlabeled sample contains only points from $S_a$, then
an active learner has to query all the points in $S_a$ to distinguish
between a hypothesis that labels all of $S_a$ positively and one that
labels positively all but one point in $S_a$. 
Since the probability of the entire set $S_b$ is $o(\epsilon)$, 
an i.i.d. sample of size $O(1/\epsilon)$, will not contain a point from $S_b$,
thus any active learner will require $\Omega(1/\epsilon)$ labels.

More formally, assume that there exists a constant $C$ and $\epsilon_0 > 0$ such that
if $\epsilon < \epsilon_0$, then at most $C/ \epsilon$ examples are
drawn. Assume from now that $\epsilon < \epsilon_0$ and that $\frac{C}
{4 \log (4/\epsilon)} \leq 1/2$. Let $A$ be the event that an i.i.d. sample
of size $m(\epsilon) \leq C/\epsilon$ contains any element from $S_b$. Then, using the
union bound, we obtain

\begin{align*}
\prob(A) \leq \frac{C}{\epsilon} \frac{\epsilon}{4 \log ( 4/
  \epsilon)} \leq 1/2 \leq 1-\delta~.
\end{align*}

% \textbf{Characterizing the Optimal Algorithm: }

\paragraph{Part 2.}
Assume now that the size of the sample is at least $\frac{ 4\log  (4/\epsilon) \log(1/(\epsilon
  \delta))}{\epsilon^2}$. It is easy to check that with probability at least
$1-\delta$, the sample contains all the points in $S_a \cup S_b$. More formally,let $\delta>0$ be any
  given confidence parameter. Let $B$ be the event that
  the sample doesn't contain all the points of $D_b$ and let $A$ the
  event that the sample doesn't contain all the points of $D_a$. For $n \in
  \{0,1,\ldots,1/\epsilon-1\}$ let $B_n$ be the event that the sample doesn't contain
  the element $b_n$. Then,

\[
\prob(B_n)  = \left(1-\frac{\epsilon^2 }{4\log (4/\epsilon)}
\right)^{\frac{ 4\log  (4/\epsilon) \log(2/(\epsilon \delta
  ))}{\epsilon^2 }} \leq \epsilon \delta /2~.
\]

Using the union bound, we obtain that 

\[
\prob(B) \leq
\frac{1}{\epsilon} \prob(A_0) \leq \delta /2~.
\]

Obviously, $\prob(A) \leq \prob(B)$. Using the union bound, we obtain that with probability at least
$1-\delta$, both $A$ and $B$ don't occur.

Given such a sample as a pool, we now show that $\OPT_{\max}  = O(\log(1/\epsilon))$,
by describing an active learning algorithm that achieves this label complexity:
\begin{enumerate}
\item For all possible separators, the points $b_0 = (1,0,0)$ and
  $b_{1/2\epsilon} = (-1,0,0)$ have different labels. The algorithm
  will first query these initial points, and then apply a binary
  search to find the boundary between negative and positive labels in
  $S_b$. This identifies the labels of all the points in $S_b$ using
  $O(\log(1/\epsilon))$ queries.

\begin{figure}
\begin{center}
\begin{tikzpicture}[scale = 1]

\draw[*-*] (-2,0) arc (180:240:2cm and 1cm);
\draw[*-*] (-2,0) arc (180:300:2cm and 1cm);
\draw[*-*] (-2,0) arc (180:360:2cm and 1cm);

\draw[-*] (-2,0) arc (180:120:2cm and 1cm);
\draw[-*] (-2,0) arc (180:60:2cm and 1cm);
\draw (-2,0) arc (180:0:2cm and 1cm);

\draw[*-*] (-1.4,2.2) arc (180:240:1.4cm and 0.7cm);
\draw[*-*] (-1.4,2.2) arc (180:300:1.4cm and 0.7cm);
\draw[*-*] (-1.4,2.2) arc (180:360:1.4cm and 0.7cm);

\draw[-*] (-1.4,2.2) arc (180:120:1.4cm and 0.7cm);
\draw[-*] (-1.4,2.2) arc (180:60:1.4cm and 0.7cm);
\draw (-1.4,2.2) arc (180:0:1.4cm and 0.7cm);

\node at ($(1.4,1)$) {$b_{n_1}$};
\node at ($(1.4,-1)$) {$b_{n_2}$};
\node at ($(2.33,0)$) {$b_{n_3}$};
\node at ($(-2.33,0)$) {$b_{n_4}$};
\node[scale=1.5]  at ($(1.2,0)$) {+};
\node[scale=1.5]  at ($(-1.2,0)$) {-};

\node at ($(1.72,2.2)$) {$a_{n_3}$};
\node at ($(-1.72,2.2)$) {$a_{n_4}$};
\node[scale=1.5] at ($(1.15,2.2)$) {+};
\node[scale=1.5]  at ($(-0.5,2.2)$) {-};

\draw[dashed] (0,-1) -- (0,1) -- cycle;
\draw[dashed] (0.9,1.7) -- (0.9,2.7) -- cycle;

\end{tikzpicture}
\end{center}
\caption{Illustration for the proof of \thmref{thm:twocircles}.}
\label{fig:mids}
\end{figure}

\item Of the points in $S_b$, half are labeled positively and half
  negatively. Moreover, there are $n_1$, $n_2$ and $y \in \{-1,1\}$
  such that $b_{n_1},\ldots,b_{n_2}$ are all labeled by $y$, and $n_2
  - n_1 + 1= |S_b|/2 = \frac{1}{2\epsilon}$ (see illustration in \figref{fig:mids}). Let $n_3 = \frac{n_2 +
    n_1}{2}$ (this is the middle point with label $y$). $n_3$ is an
  integer because $n_2-n_1$ is even, thus their sum is also even. Let
  $n_4 = \mod(n_3 + 1/2\epsilon,1/2\epsilon)$. Query the points
  $a_{n_3}$ and $a_{n_4}$ for their label. 
\item If $a_{n_3}$ and $a_{n_4}$ each have a different label, apply a binary search starting from these points 
to find the boundaries between positive and negative labels in $S_a$, using $O(\log(1/\epsilon))$ queries. Otherwise, label all the examples in $S_a$ by the label of $a_{n_3}$.
\end{enumerate}
This algorithm uses $O(\log(1/\epsilon))$ queries to label the sample. If $a_{n_3}$ and $a_{n_4}$ have different labels, it is clear that the algorithm labels all the examples correctly. We only have left to prove that if they both have the same label, then all the examples in $S_a$ also share that label. Let $h^*$ be the true hypothesis, defined by some homogeneous separator, and assume w.l.o.g that $\{ b_n \mid h^*(b_n) = 1\} = \{b_n \in S_b \mid b_n[1] > 0\}$ (note that no point has $b_n[1] = 0$ since $1/2\epsilon$ is odd).
\iffalse
cos(2pi epsilon n ) = 0 implies 2 pi epsilon n = pi or 3pi/2. In both cases we get a non-integer n.
\fi
 It follows that $n_3 = 0$ and $n_4 = 1/2\epsilon$, thus $a_{n_3} = (1/\sqrt{2}, 0, 1/\sqrt{2})$ and $a_{n_4} =
 (-1/\sqrt{2}, 0, 1/\sqrt{2})$ (see illustration in \figref{fig:mids}).  We use the following lemma, whose proof can be found in \appref{app:other}:
\begin{lemma}   \label{lem:noBinary}
Assume $1/2\epsilon$ is odd. If $\{ b_n \in S_b \mid h^*(b_n) = 1\} = \{b_n \mid b_n[1] > 0\}$
and $h^*(a_0) = h^*(a_{1/2\epsilon}) = y$ then $\forall a_n \in S_a,\quad h^* (a_n) = y$.
\end{lemma}
If follows that $\OPT_{\max} = O(\log(1/\epsilon))$.

%\textbf{Characterizing ALuMA: }
To bound the label complexity of ALuMA, it suffices to bound from below the minimal margin of 
possible separators over the given sample. 
Let $h^*$ be the correct hypothesis. 
By the same argument as in the proof of \lemref{lem:grid}, there exists some $w \in \reals^3$ that labels the sample identically to $h^*$ and attains
its maximal margin on three linearly independent points $a,b,c$
from our sample. Hence, $Aw = \mathbf{1}$ 
where $A \in \reals^{3 \times 3}$ is the matrix whose rows are $a,b,c \in S_a \cup S_b$. By Cramer's
rule, for every $i \in [3]$
\[
w[i] =  \frac{\textrm{det} A_i}{\textrm{det} A},
\]
where $A_i$ is the matrix obtained from $A$ by replacing the $i^{\text{th}}$
column  with the vector $\mathbf{1}$. Recall that the absolute value of
the determinant of $A$ is the volume
of the parallelepiped whose sides are $a,b$ and $c$. Since $a,b,c$ are 
linearly independent, each of $S_a$ and $S_b$ includes at most two of them.
Assume that $a,b \in S_a$ and $c \in S_b$. In this case, the surface area of the basis of
this parallelepiped, defined by $a$ and $b$, is at least $\frac{\sin 2 \pi \epsilon}{\sqrt{2}}$, and the height is $1/\sqrt{2}$. Hence,
\[
|\textrm{det} A| \geq \frac{\sin 2 \pi \epsilon} {2} = \Omega (\epsilon)~.
\]
The case where two of the points are in $S_b$ leads to an even larger lower bound. 
Since the elements in each $A_i$ are in $[-1,1]$,
we also have that $|\textrm{det} A_i| \leq 3! = 6$.
Thus, for $i \in [3]$ we obtain that $w_i = O(1/\epsilon)$.
All in all, we get  $\|w\|_2 = O(1/\epsilon)$, and thus 
$\gamma(h^*) = \Omega(\epsilon)~.$
% \begin{align*}
% \sin ^2 (\pi \epsilon) &= \left( \pi \epsilon - \frac{\pi^3 \epsilon^3
%   }{3!} + \frac{\pi^5 \epsilon^5
%   }{5!} - \frac{\pi^7 \epsilon^7
%   }{7!}  + \ldots \right)^2  \\ & \geq \left(\pi \epsilon -
%   \frac{\pi^3 \epsilon^3}{6} \right)^2 \\& \geq \left(\pi \epsilon -
%   \frac{\pi \epsilon}{6} \right)^2 \\ &= \frac{25}{36} \pi^2 \epsilon^2~.
% \end{align*}
Applying \corref{cor:aluma}, we obtain that ALuMA classifies all the points correctly using $O(\log(1/\gamma(h^*))\cdot \OPT_{\max}) = O(\log^2 (1/\epsilon))$ labels.
%\textbf{Characterizing CAL}
\paragraph{Part 3.}
CAL examines the examples sequentially at a random order, 
and queries the label of any point whose label is not determined by previous examples. 
Thus, if the true hypothesis is all-positive on $S_a$, and CAL sees all the points in
$S_a$ before seeing any point in $S_b$, it will request $\Omega(1/\epsilon)$ labels.
Hence, it suffices to show that there is a large probability that CAL will indeed examine all of $S_a$
before examining any point from $S_b$.
Let $A$ be the event that the first $\frac{1}{\epsilon} \log \frac{4}{\epsilon}$
examples of an i.i.d. sample contain any element from $S_b$. Then, by
the union bound, $\prob(A) \leq \frac{1}{\epsilon} \log (\frac{4}{\epsilon})\cdot \frac{\epsilon}{4 \log \frac{4}{\epsilon}} 
= 1/4$. 
Assume now that $A$ does not occur. Let $B$ be the event that the first $\frac{1}{\epsilon} \log \frac{1}{\epsilon}$ examples do not contain all the elements in $S_a$. Then, by the union
bound, $\prob(B) \leq \frac{1}{\epsilon} (1 - \epsilon )^{\frac{1}{\epsilon} \log
  \frac{4}{\epsilon}} \leq 1/4$. All in all, with probability at least $1/2$, CAL see all the points in $S_a$ before seeing any point in $S_b$ and thus its label complexity is $\Omega(1/\epsilon)$.
\end{proof}

The example above demonstrated that more unlabeled examples can help ALuMA use less labels, whereas they do not help CAL. In fact, in some cases the label complexity of CAL can be significantly worse than that of the optimal algorithm, even when both CAL and the optimal algorithm have access to all the points in the support of the distribution. This is demonstrated in the following example. Note that in this example, a passive learner also requires access to all the points in the support of the distribution, thus CAL, passive learning, and optimal active learning all require the same size of a random unlabeled pool.

\begin{ex}\label{ex:octahedron}
Consider a distribution in $\reals^d$ that is supported by two types of points
on an octahedron (see an illustration for $\reals^3$ below).
\begin{enumerate}
\item Vertices: $\{e_1,\ldots,e_d\}$. 
\item Face centers: $z/d$ for $z \in \{-1,+1\}^d$.
\end{enumerate}

Consider the hypothesis class $\W = \{ x \mapsto \sgn(\inner{x,w}-1+\frac{1}{d}) \mid w \in \{-1,+1\}^d\}.$
Each hypothesis in $\W$, defined by some $w \in \{-1,+1\}^d$, classifies at most $d+1$ data points as positive: these are the vertices $e_i$ for $i$ such that $w[i] = +1$, and the face center $w/d$. 
\end{ex}
\begin{center}
\begin{tikzpicture}[scale = 1.5]
  \draw[dashed] (1,0,0) -- (0,1,0) -- (0,0,1) -- cycle;
  \draw[dashed] (-1,0,0) -- (0,1,0) -- (0,0,1) -- cycle;
  \draw[dashed] (1,0,0) -- (0,-1,0) -- (0,0,1) -- cycle;
  \draw[dashed] (-1,0,0) -- (0,-1,0) -- (0,0,1) -- cycle;	
  \fill (0.33,0.33,0.33) circle (0.04);
  \fill (-0.33,0.33,0.33) circle (0.04);
  \fill (-0.33,-0.33,0.33) circle (0.04);
  \fill (0.33,-0.33,0.33) circle (0.04);
  \fill (1,0,0) circle (0.04);
  \fill (0,1,0) circle (0.04);
  \fill (0,0,1) circle (0.04);
  %\fill (-1,0,0) circle (0.03);
  %\fill (0,-1,0) circle (0.03);
\end{tikzpicture}
\end{center}

\begin{theorem}\label{thm:octahedron}
Consider Example \ref{ex:octahedron} for $d \geq 3$, and assume that the pool of examples includes the entire support of the distribution.
There is an efficient algorithm that finds the correct hypothesis from $\W$ with at most $d$ labels. On the other hand, with probability at least $\frac{1}{e}$ over the randomization of the sample, CAL uses at least $\frac{2^d + d}{2d+3}$ labels to find the correct separator.
\end{theorem}

\begin{proof}
First, it is easy to see that if $h^* \in \W$ is the correct hypothesis, then
\[
w = (h^*(e_1),\ldots,h^*(e_d)).
\]
Thus, it suffices to query the $d$ vertices to discover the true $w$. 

We now show that the number of queries CAL asks until finding the
correct separator is exponential in $d$. CAL inspects the unlabeled examples
sequentially, and queries any example whose label cannot be inferred from
previous labels. Consider some run of CAL
(determined by the random ordering of the sample). Assume
w.l.o.g. that each data point appears once in the sample. Let $S$ be
the set that includes the positive face center and all the
vertices. Note that CAL cannot terminate before either querying all
the $2^d-1$ negative face centers, or querying at least one example
from $S$. Moreover, CAL will query all the face centers it encounters
before encountering the first example from $S$.  At each iteration $t$
before encountering an example from $S$, there is a probability of
$\frac{d+1}{2^d+d-t}$ that the next example is from $S$. Therefore,
the probability that the first $T=\frac{2^d + d}{2d+3}$ examples are not from $S$ is
\[
\prod_{t=0}^{T-1}\left(1-\frac{d+1}{2^d+d-t}\right) \geq
\left(1-\frac{d+1}{2^d+d-T}\right)^T
\ge e^{-2T\frac{d+1}{2^d+d-T}} =  e^{\frac{-2(d+1)}{\frac{2^d+d}{T}-1}}
= \frac{1}{e} ~, 
%= \exp\left(-2\frac{d+1}{\frac{2^d+d}{T}-1}\right)^T = \exp(-1) ~,
\]
where in the second equality we used $1-a \ge \exp(-2a)$ which holds
for all $a \in [0,\half]$.
Therefore, with probability at least $\frac{1}{e}$ the number of
queries is at least $\frac{2^d + d}{2d+3}$. 
\end{proof}

These examples show that in some cases an aggressive approach is preferable to a mellow approach such as employed by CAL. At the same time, it should be noted that CAL has a guaranteed label complexity for cases for which ALuMA currently has none. Its label complexity is bounded by $\tilde{O}(d \theta\log(1/\epsilon))$, where $\theta$ is the disagreement coefficient, a
quantity that depends on the distribution and the target hypothesis
\citep{Hanneke07a,Hanneke11}. Specifically, if $D$ is uniform over a sphere
centered at the origin, then for all target hypotheses $\theta =
\Theta(\sqrt{d})$. Thus CAL achieves an exponential improvement over
passive learning for this canonical example. We do not have a similar analysis for ALuMA for the case of a uniform distribution. 
 
\iffalse
For instance, the disagreement coefficient for a uniform distribution over the points described in Example \ref{ex:dasgupta} is $1/n$. For $n = \epsilon$, the label complexity of CAL is $\tilde{\Omega}(1/\epsilon)$, similarly to the passive sample complexity. This is expected, since no active learner can avoid having to query all the points in this case.

 For instance, consider again Example
 \ref{ex:Dasgupta}, and assume that the support of the
 distribution $D$ is the set of points given in that example. As we argued above, to distinguish between the case
 in which all points have a negative label and the case in which one of
 the points has a positive label while the rest have a negative label,
 any active learning algorithm will have to query every point at least
 once. It follows that for any $\epsilon>0$, if the number of points in the support is
 at least $1/\epsilon$, then the $(\epsilon,m,D)$-label-complexity is at
 least $1/\epsilon$ for any value of $m$. On the other hand, the $\epsilon$-sample-complexity
 of passive learning in this case is $O(\frac{1}{\epsilon} \log
 \frac{1}{\epsilon})$, hence no active learner can be significantly
 better than a passive learner on this distribution. This does not contradict
 the analysis of CAL (although the example can be easily slightly changed to have a smooth distribution),
 since the disagreement coefficient here is $\Omega(1/\epsilon)$.
\fi

\subsection{Empirical Comparison}  \label{sec:experiments}

We carried out an empirical comparison between the algorithms
discussed above. Our goal is twofold: First, to evaluate ALuMA in practice, and
second, to compare the performance of aggressive strategies compared to mellow
strategies. The aggressive strategies are represented in this evaluation by ALuMA and one of the heuristics proposed by \citep{TongKoller}. 
The mellow strategy is represented by CAL. QBC represents a middle-ground between aggressive and mellow. We also compare to a passive ERM algorithm---one that uses random labeled examples.
We evaluated the algorithms over synthetic and real data sets and
compared their label complexity performance. 

% Details on our implementations and additional results are provided
% in \appref{app:experiments}.

Our implementation of ALuMA uses hit-and-run samples instead of
full-blown volume estimation, as described in \secref{sec:simpler}. QBC is also implemented using hit-and-run, as described in
\citep{KQBC}. For both ALuMA and QBC, we used a fixed number of mixing
iterations for hit-and-run, which we set to 1000.  We also fixed the
number of sampled hypotheses at each iteration of ALuMA to 1000, and
used the same set of hypotheses to calculate the majority vote for
classification. CAL and QBC examine the examples sequentially, thus
the input provided to them was a random ordering of the example pool.
The algorithm TK is the first heuristic proposed in \citep{TongKoller}, 
in which the example chosen at each iteration is the one closest
to the max-margin solution of the labeled examples known so far. 
The graphs below compare the train and the test errors of the different
algorithms. 

\begin{figure}
\centering
 \includegraphics[width = 0.4\textwidth]{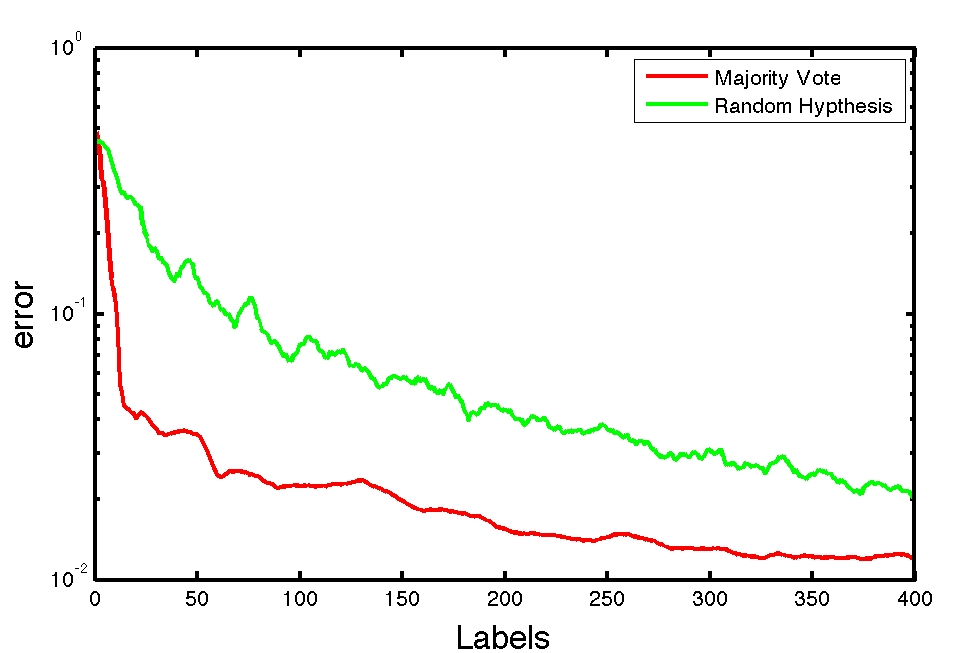}
  \caption{QBC (MNIST 4 vs. 7) - Random hypothesis Vs. Majority vote}
\label{fig:qbcRandVsMaj}
 \end{figure}

In each of the
algorithms, the classification of the training examples is done using
the version space defined by the queried labels.  The theory for CAL
and ERM allows selecting an arbitrary predictor out of the version
space. In QBC, the hypothesis should be drawn uniformly at random from
the version space. We have found that all the algorithms
show a significant improvement in classification error if they classify using the
majority vote classification proposed for ALuMA. This observation is
demonstrated in \figref{fig:qbcRandVsMaj}, which shows the rate of
error of QBC (on MNIST data which is described below) using a random hypothesis and
a majority vote. Therefore, in
all of our experiments below, the results for all the algorithms are
based on a majority vote classification.

\iffalse
\begin{table} [h!]
\begin{center}
\begin{tabular} {l | c | c | c }
Pair & \# of training examples & \# of test examples & dimension \\  \hline
$3-5$ & 11552 &  1902  & 784 \\ \hline
$4-7$ & 12107 &  2010 & 784  \\ 
\end{tabular}
\end{center}
\caption{Details of the MNIST datasets}
\end{table}
\fi

\begin{figure}
\centering
 \includegraphics[width = 0.4\textwidth]{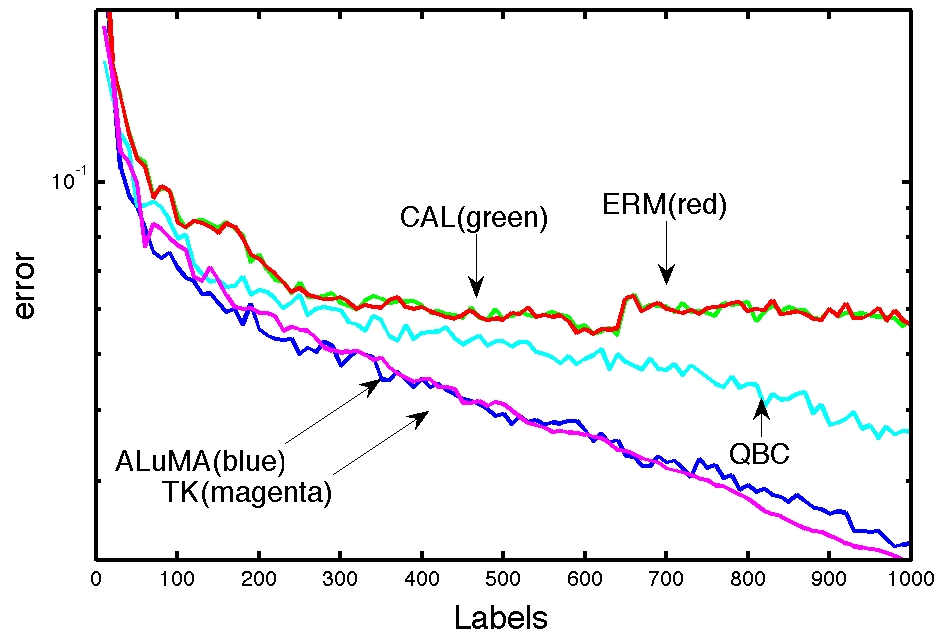}
\includegraphics[width = 0.4\textwidth]{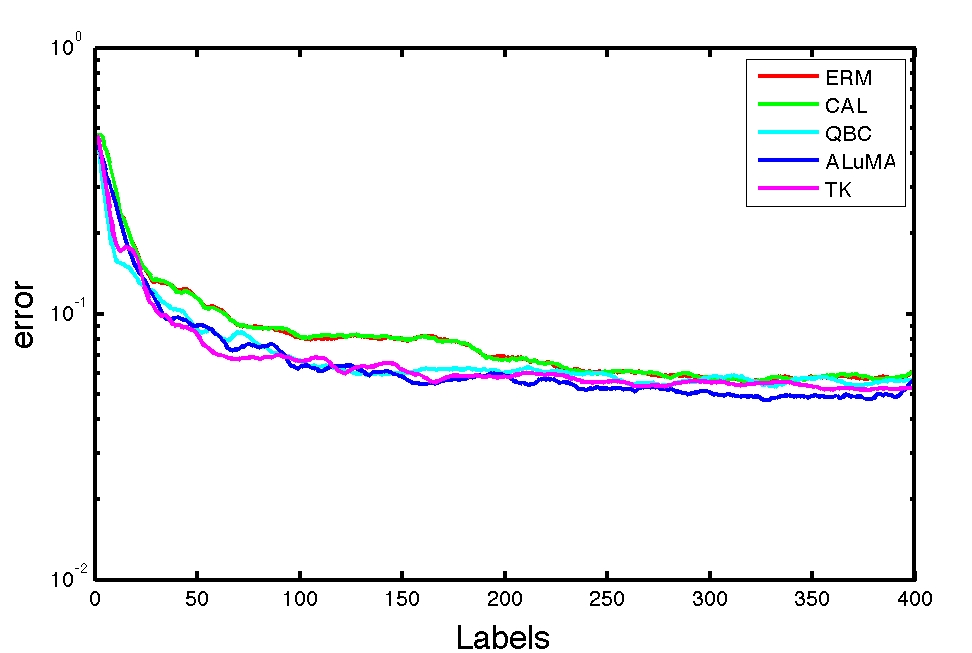}
  \caption{MNIST 3 vs. 5. Train error (left) and test error (right)}
\label{fig:mnist1}
 \end{figure}

\begin{figure}
\centering
 \includegraphics[width = 0.4\textwidth]{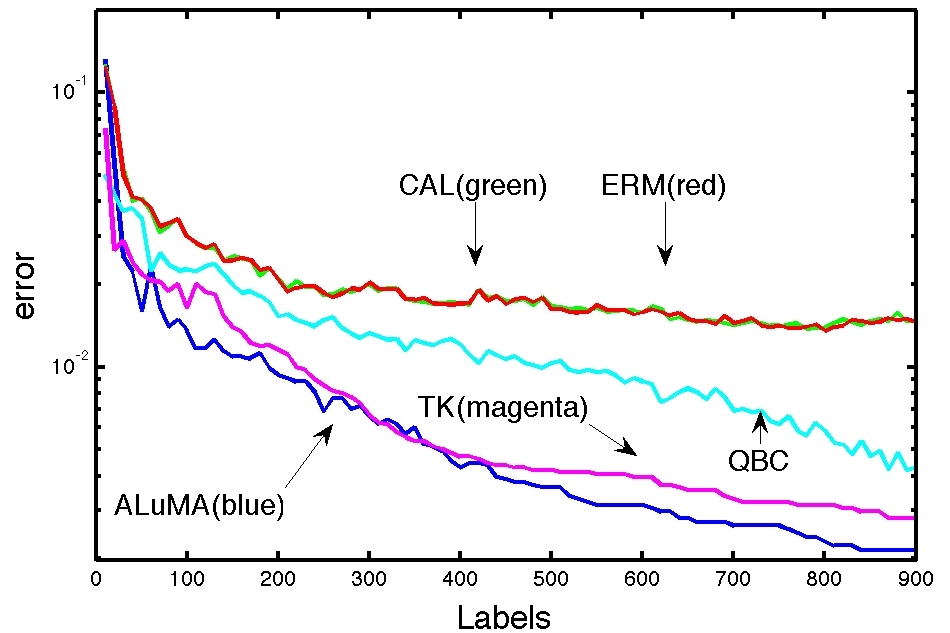} 
 \includegraphics[width = 0.4\textwidth]{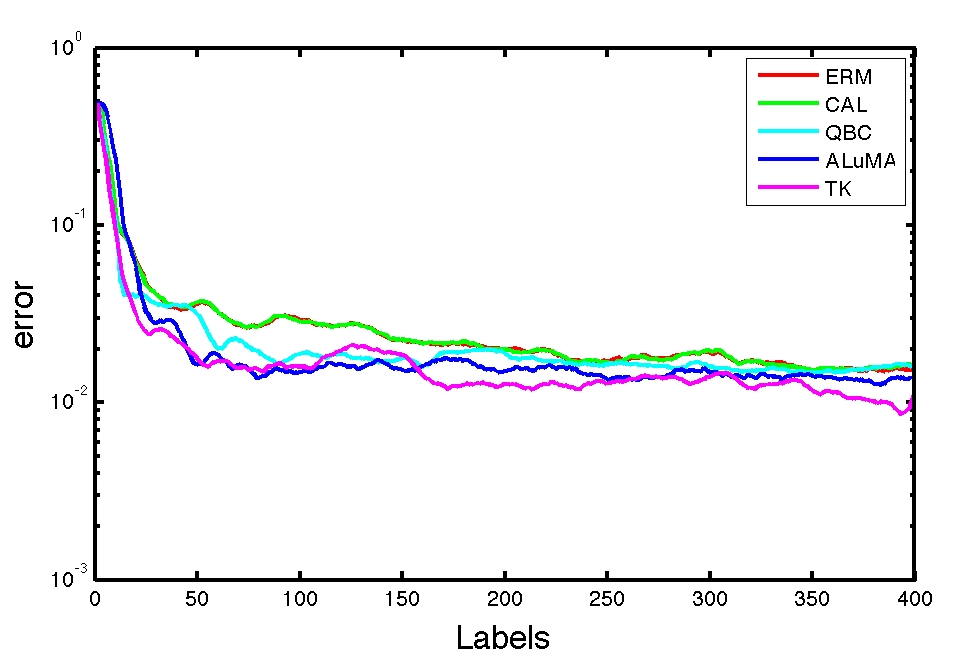}
  \caption{MNIST 4 vs. 7. Train error (left) and test error (right)}
\label{fig:mnist2}
 \end{figure}

Our first data set is MNIST\footnote{http://yann.lecun.com/exdb/mnist/}.  The examples in this
data set are gray-scale images of handwritten digits in dimension
$784$. Each digit has about $6,000$ training examples. We performed
binary active learning by pre-selecting pairs of digits. 
\figref{fig:mnist1} and \figref{fig:mnist2} depict the error as a function of the label
budget for two pairs of digits: 3 vs. 5 and 4 vs. 7. 
It is striking to observe that CAL provides no improvement over passive
ERM in the first 1000 examples, while this budget suffices to reach zero
training error for ALuMA and TK.  

\begin{figure}
%\begin{minipage}[b]{0.45\linewidth}
\centering
\includegraphics[width = 0.4\textwidth]{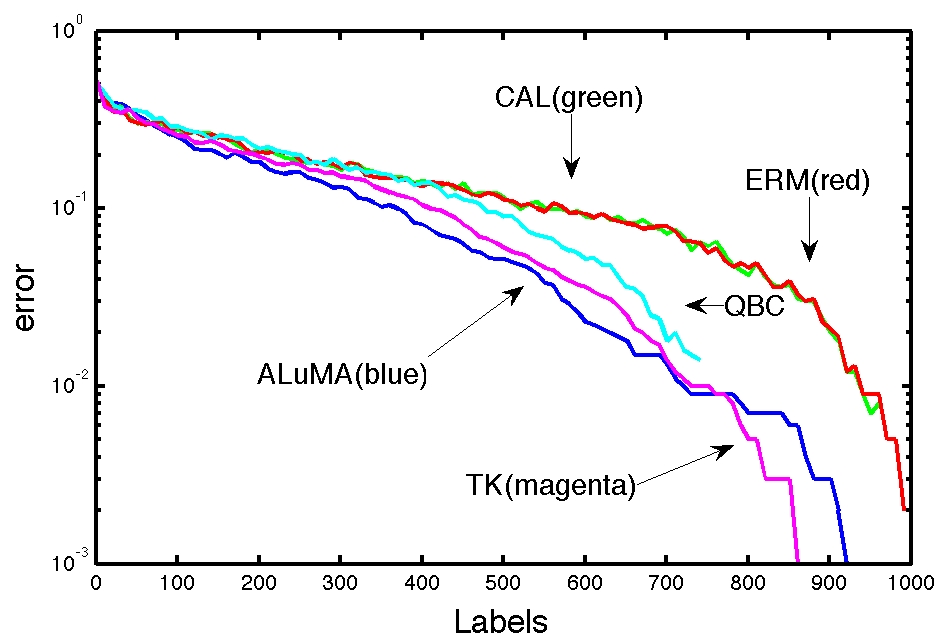}
\includegraphics[width = 0.4\textwidth]{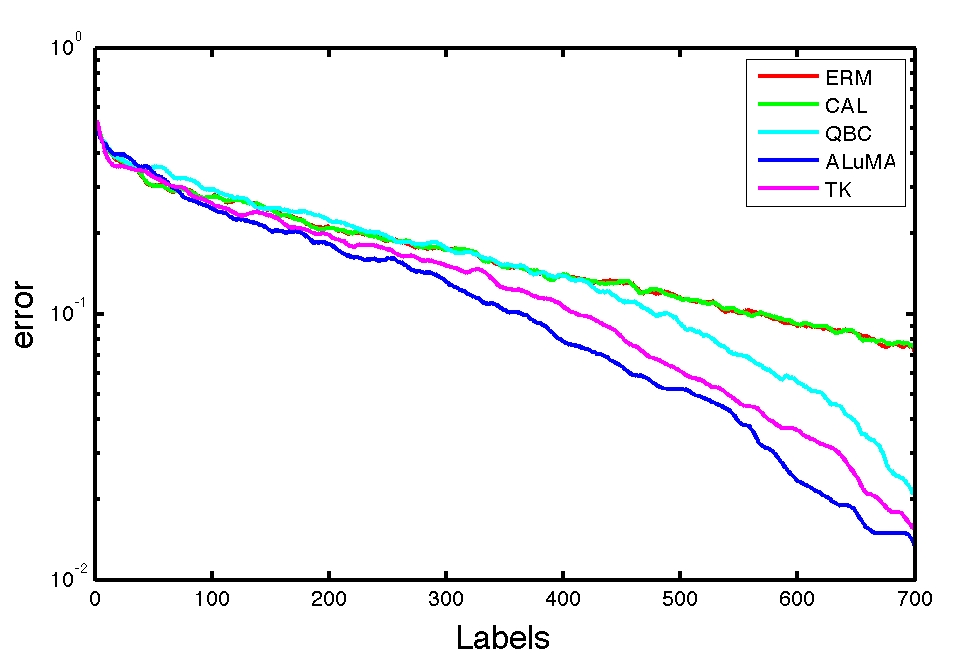}
\caption{PCMAC. Train error (left) and test error (right)}
\label{fig:mnistpcmac}
%\end{minipage}
\end{figure}

We also tested the algorithms on the PCMAC dataset\footnote{http://vikas.sindhwani.org/datasets/lskm/matlab/pcmac.mat}.  This is a
real-world data set, which represents a two-class categorization of the
20-Newsgroup collection. The examples are web-posts represented using
bag-of-words. The original dimension of
examples is $7511$. We used the Johnson-Lindenstrauss projection to reduce the
dimension to $300$, which kept the data still separable. We used a training set of $1000$ examples. \figref{fig:mnistpcmac} depicts the results. We were not able to run QBC long enough to use its entire label budget, as it tends to become slower when the training error becomes small.

\begin{figure}
%\hfill
%\begin{minipage}[b]{0.45\linewidth}
\centering
\includegraphics[width = 0.4\textwidth]{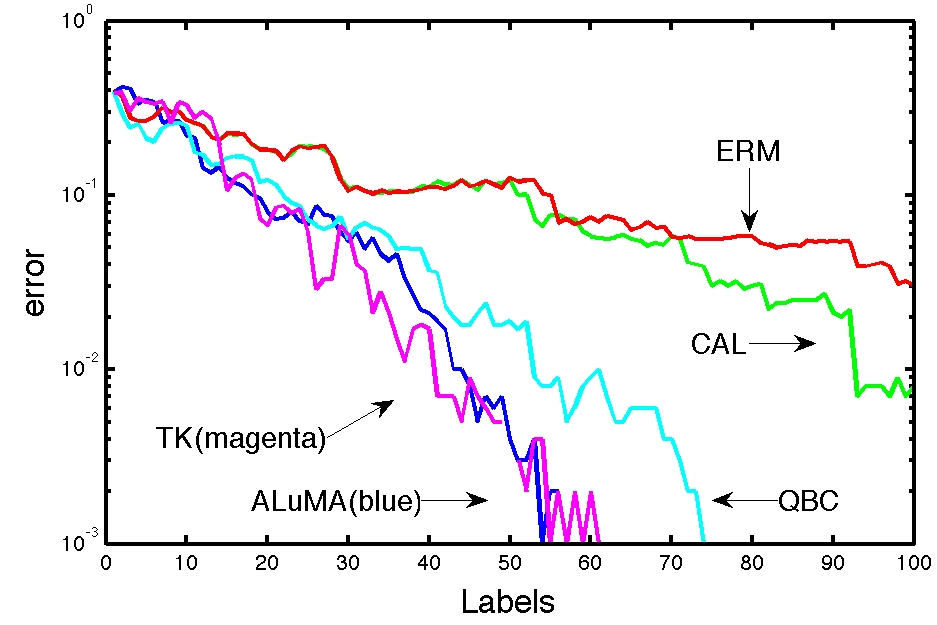}
\includegraphics[width = 0.4\textwidth]{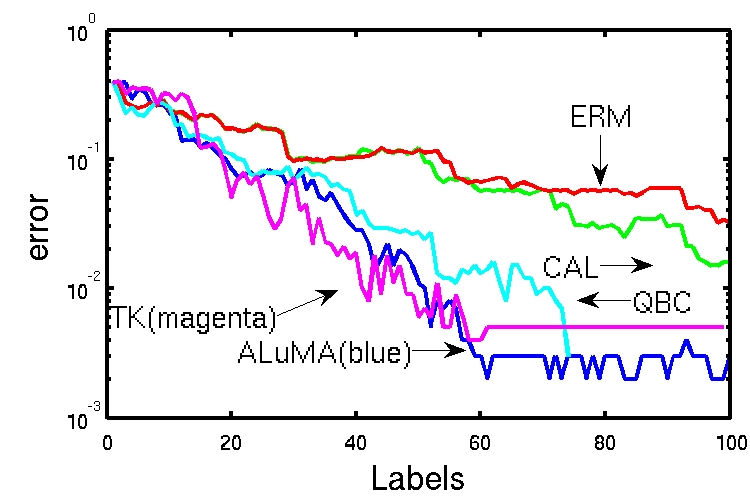}
\caption{Uniform distribution ($d = 10$). Train error (left) and test error (right)}
\label{fig:uni10}
%\end{minipage}
\end{figure}

\begin{figure}
\centering
\includegraphics[width = \textwidth]{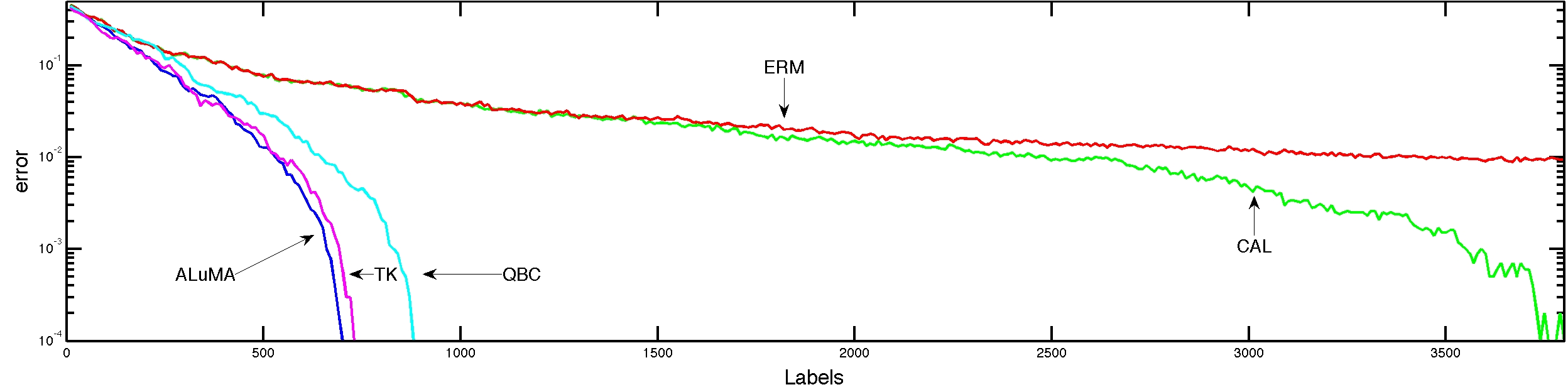}
\includegraphics[width = \textwidth]{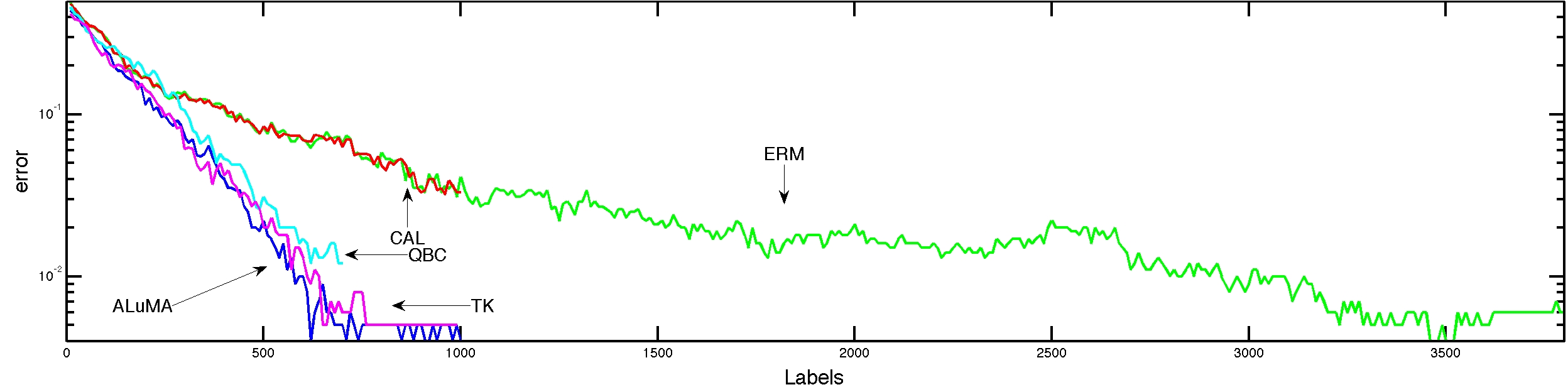}
\caption{Uniform distribution ($d = 100$). Train error (up) and test error (down)}
\label{fig:uni100}
\end{figure}

The following experiments show that ALuMA and TK outperform CAL and QBC
even on a data sampled from the uniform distribution on a sphere in
$\reals^d$. \figref{fig:uni10} and \figref{fig:uni100} depict the error as
a function of the label budget when learning a random halfspace over the
uniform distribution in $\reals^{10}$ and $\reals^{100}$ respectively. The difference between the
performance of the different algorithms is less marked for $d = 10$
than for $d=100$ , suggesting that the difference grows with the
dimension. This result suggests that ALuMA might have 
a better guarantee than the general relative analysis in the case of the
uniform distribution. Achieving such an analysis is an open question which is left for future work.

In the experiments reported so far, TK and ALuMA perform about the same, showing that the TK heuristic is very successful. However, there are cases where TK performs much worse than ALuMA, as the following synthetic experiment demonstrates. In this experiment the pool of examples is taken to be the
support of the distribution described in Example \ref{ex:octahedron},
with an additional dimension to account for halfspaces with a bias. We
also added the negative vertices $-e_i$ to the pool. 
Similarly to the proof of \thmref{thm:octahedron}, it suffices to
query the vertices to reach zero error. Table \ref{tab:octahedron} lists the number of
iterations required in practice to achieve zero error by each of the algorithms. In this experiment,
unlike the rest, ALuMA is not only much better than QBC and CAL, it is also much better than TK,
which is worse even than QBC here. 
This suggests that TK might not have guarantees similar to those of ALuMA,
despite the fact that they both attempt to minimize the same objective.
The number of queries ALuMA requires is indeed close to the
number of vertices. 

\begin{table}[h]
\centering
\begin{tabular}{|c|c|c|c|c|c|}
\hline
$d$ & ALuMA &  TK & QBC & CAL & ERM \\
\hline
$10$ & $\mathbf{29}$ & $156$ & $50$  & $308$ & $1008$ \\
$12$ & $\mathbf{38}$ & $735$ & $113$ & $862 $ & $3958$\\
$15$ & $\mathbf{55}$ & $959$ & $150$ & $2401$ & $>20000$\\
\hline
\end{tabular}
\caption{Octahedron: number of queries to achieve zero error}
\label{tab:octahedron}
\end{table}
To summarize, in all of the experiments above, aggressive algorithms performed
better than mellow ones. These results are not fully explained by current theory.
The experiments also show that ALuMA and TK have comparable success in
practice, but also that there are cases where TK is much worse than ALuMA.

\subsection{Non-separable Data}
We now turn to evaluate ALuMA on non-separable data, based on the procedure described in \secref{sec:transformation}. We compare to IWAL \citep{IWAL},
which is a state-of-the-art active
learning algorithm for the agnostic case. We compared ALuMA and IWAL to the passive soft-SVM, which selects random labeled examples from the training set as input.

\begin{figure}[h]
\centering
\includegraphics[width=0.4\columnwidth]{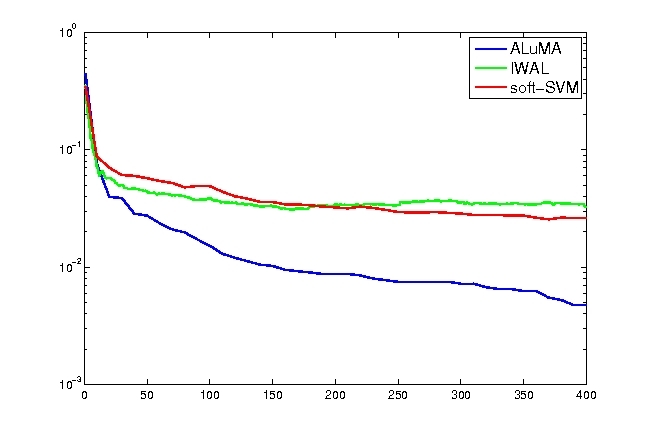}
\includegraphics[width=0.4\columnwidth]{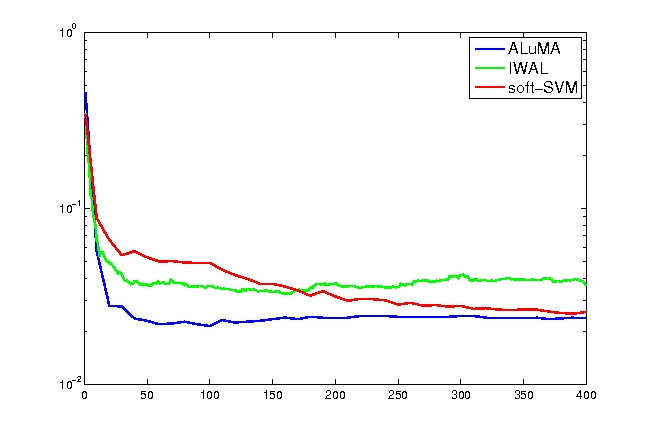}
\caption{MNIST 4
  vs. 7. (non-separable) training error (left) and test error (right)}
\label{fig:MNIST47_ag}
\end{figure}

\begin{figure}[h]
\centering
\includegraphics[width=0.4\columnwidth]{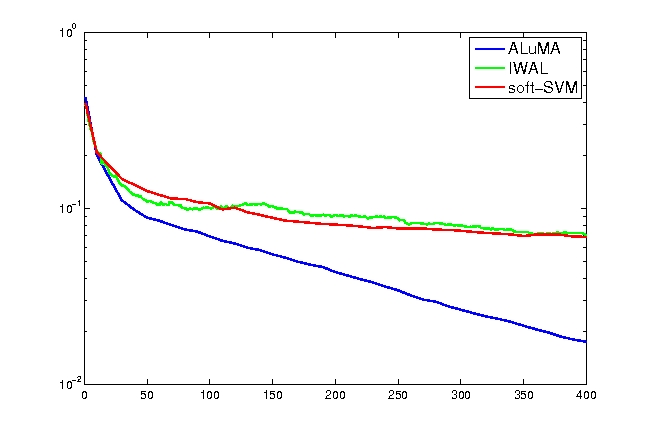}
\includegraphics[width=0.4\columnwidth]{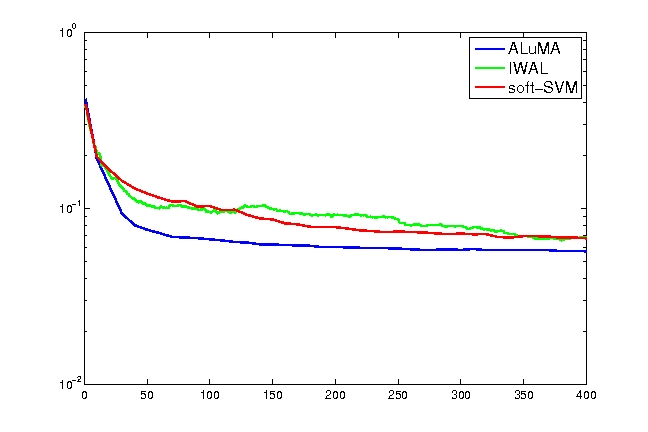}
\caption{MNIST 3 vs. 5. (non-separable), training error (left) and test error (right)}
\label{fig:MNIST35_ag}
\end{figure}

In our first experiment, we tested the algorithms on the MNIST data, pairs 3 vs. 5 and 4 vs. 7 again, by first reducing the dimension.
Following the experimental procedure in \citet{IWAL}, we projected the
$784$-dimensional data to a $25$-dimensional space using PCA. This
renders the two pairs of digits we tested in \secref{sec:experiments}
non-separable. Using model selection, we set the
regularization parameter of soft-SVM to $\lambda=10^{-3}$ and the maximal norm of the separator in IWAL to $\sqrt{1000}$. For ALuMA, the noise parameter was set to $H = 0.02$ and the dimension after preprocessing was $240$. The results
are presented in Figures \ref{fig:MNIST47_ag} and \ref{fig:MNIST35_ag}.
It can be seen that ALuMA enjoys a faster improvement in error compared to IWAL. This improvement might be attributed to the fact that we assume an upper bound on the hinge-loss in this case, while IWAL must be prepared to handle any amount of label error.

\begin{figure}[h]
\centering
\includegraphics[width = 0.4\columnwidth]{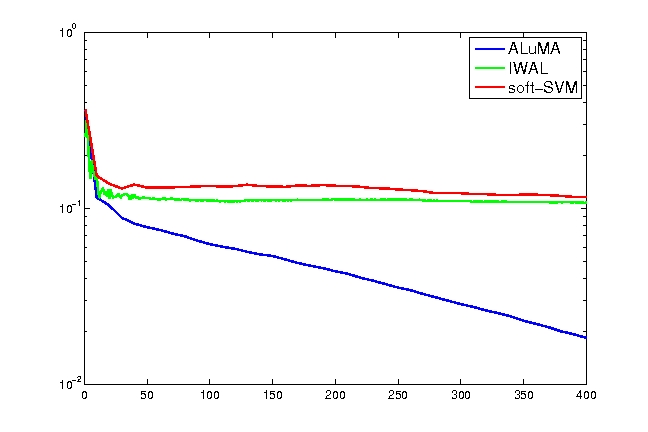}
\includegraphics[width=0.4\columnwidth]{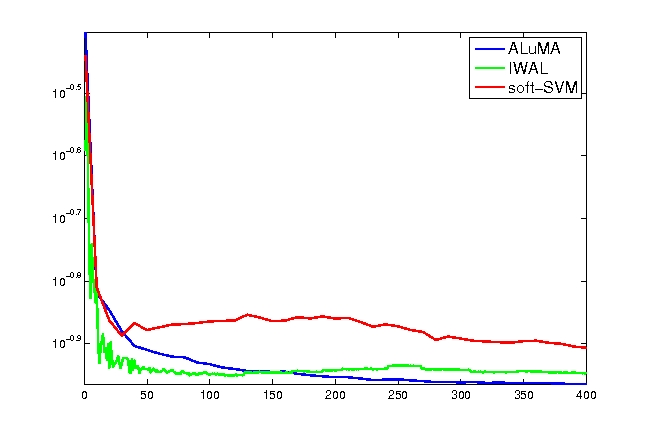}
\caption{W1A training error (left) and test error (right)}
\label{fig:w1a_ag}
\end{figure}

Our second experiment is for the  W1A data set.\footnote{http://www.csie.ntu.edu.tw/~cjlin/libsvmtools/datasets/} The original data contains
a (sparse representation of) more than $2000$ train instances and more
than $47,000$ test instances in dimension $300$. Our preprocessing step used $H = 10^{-2}$ and projected the data to dimension $260$. The other parameters were the same as in the
previous experiments. The results are shown in \figref{fig:w1a_ag}.
It can be seen that in this data set IWAL and ALuMA are comparable, both offering improvement over soft SVM. Unlike MNIST, here ALuMA does not show a consistent improvement over IWAL. We suspect that this is due to the fact that the best achievable error for this data is larger, thus decreasing ALuMA's advantage.

\section{Discussion}

In this work we have shown that the aggressive approach for active
learning can be implemented efficiently and successfully for learning
halfspaces. Our theoretical results shed light on the relationship
between the margin of the true separator and the number of active
queries that the algorithm requires. The experiments show that this approach
is practical to implement, and results in improved performance compared to mellow approaches. 

Many questions remain
open. First, while our analysis guarantees an approximation factor of
$O(d\log(m))$, in practice our experiments for the uniform distribution
show that in this case the approach performs as well or better than
algorithms which are known to achieve almost optimal rates, such as
QBC, even in high dimensions. Providing a tight analysis for the label
complexity of the aggressive approach for the uniform distribution is
thus an interesting open question. Further, while our guarantees only bound the number of queries
required to achieve zero error, in practice the algorithm performs
well compared to other algorithms even if the goal is only to reach
some small non-zero error. Characterizing the behavior of the
aggressive approach in this regime is another important open
question. Lastly, our work shows that for low-error settings, the aggressive
approach can be preferable to the mellow approach. On the other hand,
the mellow approach is clearly preferable when error levels are very
high. Thus we posit the following open problem for further research:
Characterizing the best active learning algorithm one should choose,
given a numerical upper bound on the amount of error in the given
learning problem.

%\bibliography{bib}
\bibliography{bib}

\appendix

\section{Proof of \thmref{thm:general}}  \label{sec:generalProof}
In this section we provide the complete proof of
\thmref{thm:general}. We will follow \cite{Golovin} and rely on the
notion of adaptive sub-modularity.

Denote the product space of partial realizations by $\cL_{X,\H}$. Let $f:\cL_{X,\H} \rightarrow \reals_+$ be any utility function from the set of
possible partial labelings of $X$ to the non-negative reals.
We define the notions of \emph{adaptive monotonicity} and \emph{adaptive submodularity} of a utility function using the following notation:
For an element $x \in X$, a subset $Z \subseteq X$ and a hypothesis $h \in \H$, we define the conditional expected marginal benefit of $x$,
conditioned on having observed the partial labeling $h|_{Z}$, by
\[
\Delta(h|_Z,x) = \E _{g} \big[f(g|_{Z\cup \{x\}})-f(g|_Z)\:\big|\: g|_Z =  h|_{Z}\big].
\]
Put another way, $\Delta(h|_Z,x)$ is the expected improvement of $f$
if we add to $Z$ the element $x$, where expectation is over a choice
of a hypothesis $g$ taken uniformly at random from the set of
hypotheses that agree with $h$ on $Z$. 

\begin{definition}[Adaptive Monotonicity] \label{def:ad_mon} 
A utility function $f: \cL_{X,\H} \rightarrow \reals_+$ is \emph{adaptive monotone} if the conditional expected marginal benefit is always non-negative. That is, if for all $h \in \H, Z \subseteq X$ and $x \in X$, $\Delta(h|_Z,x) \geq 0$.
\end{definition}
 
\begin{definition}[Adaptive Submodularity] \label{def:ad_sub} 
 A function $f: \cL_{X,\H} \rightarrow \reals_+$ is \emph{adaptive submodular}
 if the conditional expected marginal benefit of a given item does not increase
if the partial labeling is extended. That is, if for all $h \in \H$, for
 all $Z_1 \subseteq Z_2 \subseteq X$ ,and for all $x \in X$,
 \[
 \Delta(h|_{Z_1},x) \geq \Delta(h|_{Z_2},x).
 \]  
 \end{definition}
 
Any (deterministic) pool-based algorithm is associated with
  a policy
function, which we usually denote by $\pi$, which maps each partial realization $h|_S$ to an
element $x$ of
$X$, namely, the element $x$ queried by the algorithm after observing
$h|_S$. It is natural to consider a greedy algorithm which always selects
an item that maximizes the marginal utility. Since it is often computationally hard to choose the element which maximizes the
marginal utility, we introduce the notion of an approximately-greedy
algorithm, following \citet{Golovin}.

\begin{definition}[Approximate Greedy]
Let $\alpha \geq 1$. An algorithm which is associated with policy $\pi:\cL_{X,\H} \rightarrow X$ is \emph{$\alpha$-approximately greedy}
with respect to a utility function $f$ if for every $h$ and for every $Z \subseteq X$
\begin{equation} \label{eq:alpha}
\Delta(h|_Z,\pi(h|_Z)) \geq \frac{1}{\alpha} \max_{x \in X} \Delta(h|_Z,x).
\end{equation}

If an algorithm $\cA$ is \emph{$1$-approximately greedy}
with respect to a utility function $f$, we simply say that $\cA$ is greedy
w.r.t. $f$.
\end{definition} 

We denote by $S(\cA, h, k)$ the first $k$
pairs of instances along with their labels observed by $\cA$, under the assumption that $L \determined
h$. Following this notation, the utility of running $\cA$ for $k$
 steps under the assumption that $L \determined h$ is denoted by $f(S(\cA,h,k))$. The expected utility of
 running $\cA$ for $k$ steps is defined by 

\[
f_\avg (\cA, k) = \E_{h \sim P} [f(S(\cA,h,k)].
\]

The central theorem of adaptive submodularity, stated below as \thmref{thm:ad_Max},
 links the expected utility of the optimal policy for maximizing
 $\favg$ with the expected utility of the \fix{associated approximately-greedy
 algorithm}. 

\begin{theorem}[\cite{Golovin}] \label{thm:ad_Max}
Let $f: \cL_{X,\H} \rightarrow \reals_+$ be a utility
function, and let $\cA$ be a (deterministic) active learning algorithm. If $f$ is adaptive
monotone and adaptive submodular, and $\cA$ is $\alpha$-approximately
greedy, then for any deterministic algorithm $\cA^*$ and for all positive integers $t, k$,
\begin{equation}\label{eq:avgapprox}
\favg(\cA, t) \geq (1-e^{-\frac{t}{\alpha k}}) \favg (\cA^*, k). 
\end{equation}
\end{theorem}

Let $P$ be a distribution over $\H$. For any algorithm $\alg$, denote by $V_t(\alg,h)$ the
version space induced by the first $t$ labels it queries if the true
labeling of the pool is consistent with $h$. Denote the version space
reduction of $\alg$ after $t$ queries in the case that $L \Leftarrow
h$ by

\begin{equation}   \label{eq:utilityfh}
f(\alg, t, h) = 1-P(V_t(\alg, h)).
\end{equation}

The average version space reduction of $\alg$ after $t$ queries is 
\[
f_\avg(\alg, t) = 1-\E_{h \sim P}[P(V_t(\alg, h))].
\]

In the active learning setting, we define the utility function $f$ as in \eqref{eq:utilityfh}
and have the following result:
\begin{lemma}[\cite{Golovin}] \label{lm:f}
The function $f$ defined in \eqref{eq:utilityfh} is adaptive monotone and adaptive submodular.
\end{lemma}

%Therefore \thmref{thm:ad_Max} holds for this utility function. 

\begin{corollary}   \label{cor:average}
Let $X=\{x_1, \ldots, x_m\}$. Let $\H$ be a hypothesis class, and let $P$ be a
  distribution over $\H$. Suppose that $\cA$ is $\alpha$-approximately
  greedy with respect to $P$, and let $\cA^*$ be \fix{a (deterministic) algorithm} that achieves $\OPT_{\max}$, that is $c_{wc}(\cA^*) = \OPT_{\max}$. Then, for all positive integers $t, k$,

\[
\favg(\cA, t) \geq (1-e^{-\frac{t}{\alpha k}}) \favg (\cA^*, k). 
\]
\end{corollary}

The following lemma will allow us to show that the version space of an $\alpha$-approximately greedy algorithm is relatively pure.
\begin{lemma}\label{lem:multvols}
Let $\cA^*$ be an algorithm that achieves $\OPT_{\max}$.
For any $h \in \H$, any active learner $\cA$, 
and any $t$,
\[
\favg(\cA^*,\OPT_{\max}) - \favg(\cA,t) \ge
P(V(h|_X))\left(P(V_t(\cA,h)) - P(V(h|_X)) \right) ~.
\]
\end{lemma}
\begin{proof}
Since $\cA^*$ acheives the optimal worst-case cost, the version space induced by the labels that $\cA^*$ queries within the first $\OPT_{\max}$ iterations must be exactly the set of hypotheses which are consistent with the true labels of the sample. Therefore, for any $h \in \H$. 
\[
P(V_{\OPT_{\max}}(\cA^*,h)) = P(V(h|_X)).
\]
By definition of $\favg$,
\begin{align*}
\favg(\cA^*, \OPT_{\max})-\favg(\cA,t) &= \E_{h \sim P}[P(V_t(\cA,h)) - P(V_{\OPT_{\max}}(\cA^*,h))]\\
&= \E_{h \sim P}[P(V_t(\cA,h)) - P(V(h|_X))].
\end{align*}
Since $S(\cA,h,t)$ does not depend on the value of $h$ outside of $X$, we can sum over the possible labelings of $X$ to have
\begin{align*}
\favg(\cA^*, \OPT_{\max})-\favg(\cA,t) &=\sum_{h|_X: h \in \H} P(V(h|_X)) (P(V_t(\cA,h)) - P(V(h|_X))).
\end{align*}
Now, it is easy to see that for any $h \in \H$, $V_t(\cA,h) \supseteq V(h|_X)$, thus \[
P(V_t(\cA,h)) - P(V(h|_X)) \geq 0.
\]
It follows that for any $h \in \H$
\[
\favg(\cA^*, \OPT_{\max})-\favg(\cA,t) \geq 
P(V(h|_X)) (P(V_t(\cA,h)) - P(V(h|_X))).
\]
\end{proof}

Combining \corref{cor:average} and \lemref{lem:multvols}, the following corollary is immediate.
\begin{corollary}   \label{cor:quotientVol}
For any $\alpha$-approximate greedy algorithm $\cA$, 
\[
\forall h\in \H,\quad
P(V(h|_X)) (P(V_t(\cA,h)) - P(V(h|_X))) ~\le~
e^{-\tfrac{t}{\alpha\OPT_{\max}}} ~,
\]
which yields
\begin{equation} \label{eq:pobefmar2}
\forall h\in \H,\quad
\frac{P(V(h|_X))}{P(V_t(\cA,h))}  \geq  \frac{P(V(h|_X))^2}{e^{-\tfrac{t}{\alpha\OPT_{\max}}} + P(V(h|_X))^2}.
\end{equation}
\end{corollary}

\begin{proof}  \textbf{(Of \thmref{thm:general})}
Let $\cA$ be $\alpha$-approximately
greedy algorithm which outputs a $\beta$-approximate majority vote. \corref{cor:quotientVol} holds for $\cA$. Let $h$ be the
target hypothesis. Substituting $T
\geq \alpha(2\ln(1/P(h)) +
\ln(\frac{\beta}{1-\beta}))\cdot\OPT_{\max}$ into \eqref{eq:pobefmar2}
implies that 

\[
\frac{P(V(h|_X))}{P(V_T(\cA,h))}  \geq \beta~.
\]

The proof now follows from the fact that $\cA$ outputs a $\beta$-approximate majority vote.
\end{proof} 

% This means that if $P(V(h|_X))$ is large enough and we run an
% approximate greedy policy, then after a sufficient number of
% iterations, most of the remaining version space induces the correct
% labeling of the sample. 

\section{Handling Non-Separable Data and Kernel Representations} \label{app:transformation}

We now prove \thmref{thm:transform} by showing that \algref{algo:preprocess}
satisfies the claims of the theorem.  It is clear that \algref{algo:preprocess}
is polynomial as required in item (\ref{item:poly2}). In addition, item (\ref{item:dim}) holds from the
definition of \algref{algo:preprocess}. We have left to prove item (\ref{item:separable}). We first prove that it holds for
the case where the input is represented directly as $X \subseteq \reals^d$.

We start by showing that under the assumption of \thmref{thm:transform}, the set $\{x'_1,\ldots,x'_m\}$, which is generated in step \ref{step:setxprime},
is separated with a bounded margin by the original labels of $x_i$.
Fix $\gamma > 0$ and $w^* \in \ball^d$.
For each $i \in [m]$, define 
\[
\ell_i = \max(0,\gamma-L(i) \inner{w^*, x_i}).
\]
Thus, $\ell_i$ quantifies the margin violation of example $x_i$ by $w^*$, relative to its true label $L(i)$.

\begin{lemma}
If $H \geq \sum_{i=1}^m\ell^2_i$, where $H$ is the input to \algref{algo:preprocess},
then there is a $w \in \ball^{d+m}$ such that for all $i\in [m]$, $L(i)\inner{w,x'_i} \geq  \frac{\gamma}{1+\sqrt{H}}$.
\end{lemma}
\begin{proof}
By step \ref{step:setxprime} in \algref{algo:preprocess}, 
$x'_i = (a\cdot x_i; \sqrt{1-a^2}\cdot e_i)$, where $a = \sqrt{\frac{1}{1+\sqrt{H}}}$. 
Define
\[
w' = (w^*; \frac{a}{\sqrt{1-a^2}}(L(1)\ell_1,\ldots,L(m)\ell_m)).
\]

Then 
\[
L(i)\inner{w', x'_i} = a L(i)\inner{w^*,x_i} + a\ell_i \geq a(\gamma -\ell_i) + a\ell_i = a\gamma. 
\]

Let $w = \frac{w'}{\|w'\|}$. Then $w \in \ball^{d+m}$, and 
\[
L(i)\inner{w, x'_i} = \frac{L(i)\inner{w', x'_i}}{\|w'\|} \geq \frac{a\gamma}{\sqrt{1 + \frac{a^2}{1-a^2}\sum_{i=1}^m\ell^2_i}} = \frac{\gamma}{\sqrt{\frac{1}{a^2} + \frac{1}{1-a^2}\sum_{i=1}^m\ell^2_i}}.
\]
Set $a^2 = \frac{1}{1+\sqrt{H}}$, and assume $H \geq \sum_{i=1}^m\ell^2_i$.
Then 
\[
L(i)\inner{w, x'_i} \geq \frac{\gamma}{1+\sqrt{H}}.
\]
\end{proof}

The set $\{\bar{x}_1,\ldots,\bar{x}_m\}$ returned by \algref{algo:preprocess}
is a Johnson-Lindenstrauss projection of $\{x'_1,\ldots,x'_m\}$ on $\reals^k$.
It is known \citep[see e.g.][]{Blum} that if a set of $m$ points is separable with margin $\eta$ and
$k \geq O\left(\frac{\ln(m/\delta)}{\eta^2}\right)$, then with  probability $1-\delta$,
the projected points are separable with margin $\eta/2$. Setting $\eta = \frac{\gamma}{1+\sqrt{H}}$,
it is easy to see that step \ref{step:jlproj} in \algref{algo:preprocess} indeed maintains the desired margin. This completes the proof of item (\ref{item:separable}) of \thmref{thm:transform} for the case where the input is $X \subseteq \reals^m$.

We now show that if the input is a kernel matrix $K$, then 
the decomposition step \ref{step:decompose}
preserves the separation properties of the input data, thus showing that
item (\ref{item:separable}) holds in this case as well. To show that our decomposition step
does not change the properties of the original data, we first
use the following lemma, which indicates that separation properties are conserved under different decompositions of the same kernel matrix.

\begin{lemma}[\cite{SabatoSrebroTishby}, Lemma 6.3]\label{lem:sab}
Let $K \in \reals^{m \times m}$ be a positive definite matrix and let $V \in \reals^{m\times n}, U\in \reals^{m \times k}$ be matrices such that $K = VV^T = UU^T$. 
For any vector $w \in \reals^n$ there exists a vector $u \in \reals^k$ such that $V w = U u$ and $\|u\| \leq \| w\|$.
\end{lemma}

The next lemma extends the above result, showing that the property holds even if $K$ is not invertible.
\begin{lemma}\label{lem:extend}
Let $K \in \reals^{m \times m}$ be a positive definite matrix and let $V \in \reals^{m\times n}, U\in \reals^{m \times k}$ be matrices such that $K = VV^T = UU^T$. 
For any vector $w \in \reals^n$ there exists a vector $u \in \reals^k$ such that $V w = U u$ and $\|u\| \leq \| w\|$.
\end{lemma}
\begin{proof}
For a matrix $A$ and sets of indexes $I,J$ let $A[I]$ be the sub-matrix of $A$
whose rows are the rows of $A$ with an index in $I$. Let $A[I,I]$ be the sub-matrix of $A$ whose rows and columns are those that have index $I$ in $A$.

If $K$ is invertible, the claim holds by \lemref{lem:sab}. Thus, assume $K$ is
singular.  Let $I \subseteq [m]$ be a maximal subset such that the matrix
$K[I;I]$ is invertible.\footnote{if no such subset exists then $K,V,U$ are all zero and the claim is trivial.} Then by \lemref{lem:sab}, $K[I;I] = V[I] (V[I])^T =
U[I] (U[I])^T$, and there exists a vector $u$ such that $V[I] w = U[I] u$, and
$\|u\| \leq \| w\|$. We will show that for any $i \notin I$, $V[i] w = U[i] u$
as well.

For any $i \notin I$, $K[I\cup\{i\};I\cup\{i\}]$ is singular. 
Therefore $V[I\cup\{i\}]$ is singular, while $V[I]$ is not.
Thus there is some vector $\lambda \in \reals^{|I|}$ such that 
$V[i]^T = V[I]^T \lambda$. By a similar argument there is some vector $\eta \in \reals^{|I|}$ such that $U[i]^T = U[I]^T\eta$. We have $K[I,i] = V[I]V[i]^T = V[I]V[I]^T \lambda = K[I,I]\lambda$. Similarly for $U$, $K[I,i] = K[I,I]\eta$.
Therefore $K[I,I](\lambda - \eta) = 0$. Since $K[I,I]$ is invertible, 
it follows that $\lambda = \eta$. Therefore, $U[i] u = \eta^T U[I] u = \lambda^T V[I] w = V[i] w$. 
\end{proof}

We now use this lemma to show that the decomposition step does not 
change the upper bound on the margin loss which is assumed in \thmref{thm:transform}.

\begin{theorem}
Let $\psi_1,\ldots,\psi_m$ be a set of vectors in a Hilbert space $S$,
and let $K \in \reals^{m\times m}$
such that for all $i,j \in [m]$, $K_{i,j} = \inner{\psi_i,\psi_j}$.
suppose there exists a $w \in S$ with $\|w\| \leq 1$ such that 
\begin{equation}\label{eq:h1}
H \geq \sum_{i=1}^m \max(0,\gamma -  y_i\inner{w, \psi_i})^2.
\end{equation}

Let $U \in \reals^{m \times k}$ such that $K = U U^T$ and let $x_i$ be
row $i$ of $U$.  Then there exists a $u \in \ball^k$ such that 
\begin{equation}\label{eq:h2}
H \geq \sum_{i=1}^m \max(0,\gamma -  y_i\inner{u, x_i})^2.
\end{equation}
\end{theorem}
\begin{proof}
Let $\alpha_1,\ldots,\alpha_n \in S$ be an orthogonal basis for the span of
$\psi_1,\ldots,\psi_m$ and $w$, and let $v_1,\ldots,v_m, v_w \in \reals^n$ such
that $\sum_{l=1}^n v_i[l] \alpha_l = \psi_i$ and $\sum_{l=1}^n v_w[l] \alpha_l
= w$.  Let $V \in \reals^{m\times n}$ be a matrix such that row $i$ of the
matrix is $v_i$. Then $K = V V^T$, and $V v_w = r$, where $r[i] = \inner{v_w,
  v_i} = \inner{w, \psi_i}$. By
\lemref{lem:extend}, there exists a $u \in \reals^k$ such that $U u = r$. Then we have
$\inner{u, x_i} = r[i]$. Therefore for all $i \in [m]$, $\inner{w, \psi_i} =
\inner{u, x_i}$, thus \eqref{eq:h1} implies \eqref{eq:h2}. In addition, $\|u\|
\leq \|v_w\| = \| w\| \leq 1$, therefore $u \in \ball^k$.
\end{proof}

\section{Other Proofs}\label{app:other}

In this section we provide proofs omitted from the text. 

\begin{proof}[of \lemref{lem:phbound}]
Fix $h\in\cW$ and let $V = \{h' \in \H : \forall i, h'(x_i)=h(x_i)\}$. Assume w.l.o.g.\ that $\|x\| = 1$ for all $x \in X$. Denote for brevity $\gamma = \gamma(h)$.
Choose $w \in \ball^d$ such that $\forall x \in X,\, h(x)\inner{w,x}\geq \gamma$.
For a given $v \in \ball^d$, denote by $h_{v} \in \H$ the mapping $x \mapsto \sgn(\inner{v,x})$. Note that for all $v \in \ball^d$ such that $\|w-v\| < \gamma$, $h_v \in V$. This is because for all $x \in X$,
\begin{align*}
&h(x)\inner{v,x} = \inner{v-w, h(x)\cdot x} + h(x)\inner{w, x} \geq-\|w - v\|\cdot\|h(x) \cdot x\| + \gamma > -\gamma + \gamma = 0,
\end{align*}
Since $h_v(x) = \sgn(\inner{v,x})$ it follows that $h_v(x) = h(x)$.
We conclude that $\{v \mid h_{v} \in V\} \supseteq \ball^d \cap B(w,\gamma)$,
where $B(z,r)$ denotes the ball of radius $r$ with center at $z$.
Let $u = (1-\gamma/2)w$. Then for any $z \in B(u,\gamma/2)$, we have $z \in \ball^d$, 
since \[
\|z\| = \|z-u+u\| \leq \|z-u\| + \|u\| \leq \gamma/2 + 1-\gamma/2 = 1.
\]
In addition, $z \in B(w,\gamma)$ since
\[
\|z-w\| = \|z-u+u-w\| \leq \|z-u\| + \|u-w\| \leq \gamma/2 + \gamma/2 = \gamma.
\] 
Therefore $B(u,\gamma/2) \subseteq \ball^d \cap B(w, \gamma).$
We conclude that $\{v \mid h_{v} \in V\} \supseteq B(u,\gamma/2)$. Thus,
\[
P(h) = P(V) \geq \Vol(B(u,\gamma/2))/\Vol(\ball^d) \geq \left(\frac{\gamma}{2} \right)^d.
\]
\end{proof}

\begin{proof} \textbf{(of \lemref{lem:grid})}
Let us multiply all examples in the pool by $1/c$. Then, all the
elements of all examples in the pool are integers. Choose a labeling $L$
which is consistent with some $w^*$. Consider the optimization problem:
\[
\min_{w} \|w\|^2 ~~\textrm{s.t.}~~ \forall i,~ L(i)\inner{w,x_i} \ge 1 ~.
\]
For simplicity assume that the pool of examples span all of
$\reals^d$.  Then, it is easy to show that if $w$ the solution to the
above problem then there exist $d$ linearly independent examples from
the pool, denoted w.l.o.g. by $x_1,\ldots,x_d$, such that $L(i)
\inner{w,x_i} = 1$ for all $i$. In other words, $w$ is the solution of
the linear system $Aw=b$ where the rows of $A$ are $x_1,\ldots,x_d$
and $b = (L(1),\ldots,L(m))^T$. 

\renewcommand{\det}{\textrm{det}}

By Cramer's rule, $w_i = \det(A_i)/\det(A)$, where $A_i$ is obtained
by replacing column $i$ of $A$ by the vector $b$. Since all elements
of $A$ are integers and $A$ is invertible, we must have that $|\det(A)| \ge
1$.  Therefore, $|w_i| \le |\det(A_i)|$. Furthermore, by Hadamard's
inequality, $|\det(A_i)|$ is upper bounded by the product of the norms
of the columns of $A_i$. Since each element of $A_i$ is upper bounded
by $1/c$, we obtain that the norm of each column is at most
$\tfrac{\sqrt{d}}{c}$, hence $|\det(A_i)| \le (\sqrt{d}/c)^d$.  It
follows that $\|w\| \le \sqrt{d}\,(\sqrt{d}/c)^d$. Hence, the margin
is
\[
\frac{1}{\|w\|\,\max_i \|x_i\|} \ge
\frac{1}{\sqrt{d}\,(\sqrt{d}/c)^d\,\cdot\,\sqrt{d}/c}
= \frac{1}{\sqrt{d}\,(\sqrt{d}/c)^{d+1}} ~.
\]

\end{proof}

\begin{proof}  \textbf{(of \thmref{thm:lower-bound})}
Set $m = \lfloor \ln(1/\gamma)\rfloor$ such that $m$ is a power of $2$. Let
$x'_0=(1,0) \in \reals^2$. For all $i \in [m-1]$, define $x'_i = (\cos(\pi/2^i), \sin (\pi /2^i))$. Fix $c > 0$, and define $S = \ball^2 \cap \{-1,-1+c, \ldots,
1-c,1\}^2$.
For each $i \in \{0, 1, \ldots ,m-1\}$, let $x_i$ be the nearest
neighbor of $x'_i$ in $S$, that is $x_i = \arg \min_{x \in S} \|x-x'_i\|_2$.
It can be easily seen that if $c = \Theta(\gamma)$ then $\forall i,\: \|x_i - x'_i\| = O(\gamma)$. 

Consider an exact greedy algorithm that always selects $x_0$ first (this is possible since on the first round of the algorithm, any query halves the version space). Suppose that the target hypothesis $h^*$ satisfies 
\[
h^*(x_i) = \begin{cases} -1 &  i=0 \\ 1 & \text{otherwise}   \end{cases}
\]
By setting a small enough $c$ we get that $\gamma(h^*) = \Omega(\gamma).$ 

If $c$ is small enough compared to $\gamma$, then after querying $x_0$ the algorithm will query $x_1, \ldots, x_{m-1}$ in order. In addition, on every round $t < m-1$ the majority vote would lead to the wrong labeling, since only a small fraction of the version space belongs to the correct hypothesis. Thus the algorithm queries all the examples (except perhaps one) before reaching the correct answer.
\end{proof}

\begin{proof}[of \lemref{lem:noBinary}]
We prove the lemma for the case $h^*(a_{1/2\epsilon}) = 1$. The case $h^*(a_0) = -1$ can be proved similarly.
Let $w^*$ be any hyperplane which is consistent with $h^*$. Let $n_1 =
\frac{1}{4 \epsilon}- \half$ and let $n_2 = n_1+1$. Then 
\begin{align*}
b_{n_1} &= (\cos(\pi/2-\pi \epsilon), \sin(\pi/2-\pi \epsilon), 0),\text{ and }\\
b_{n_2} &= (\cos(\pi/2+\pi \epsilon), \sin(\pi/2+\pi \epsilon), 0).
\end{align*}
By the assumption of the lemma, $\inner{w^*,b_{n_1} } > 0$ and $\inner{w^* ,b_{n_2} } < 0$.
It follows that $w^*[1] \sin \pi \epsilon > w^*[2] \cos \pi \epsilon$ and $-w^*[1] \sin \pi \epsilon < w^*[2] \cos \pi \epsilon$. As a consequence, we obtain that $|w^*[2]| < w^*[1] \tan(\pi \epsilon)$.

Now, choose some $n \in \{0, \ldots, 1/\epsilon -1\}$. We show that
the corresponding element in $S_a$ is labeled positively. First, from
the last inequality, we obtain
\begin{align}  \label{eq:geq1}
\inner{w^*, a_n} &= \frac{1}{\sqrt{2}}\inner{w^* , (\cos 2 \pi \epsilon n , \sin
2 \pi \epsilon n,1) } \notag\\
&\geq \frac{1}{\sqrt{2}}(w^*_1
(\cos 2 \pi \epsilon n - \tan(\pi \epsilon) \sin (2 \pi \epsilon n))
+ w^*_3).
\end{align}
We will now show that 
\begin{equation}  \label{eq:geq2}
\forall n \in \{0,1,\ldots, 1/\epsilon-1\},~
\cos 2 \pi \epsilon n - \tan(\pi \epsilon) \sin (2 \pi
\epsilon n) \geq -1.
\end{equation}

From symmetry, it suffices to prove this for every $n \in \{0,1, \ldots, 1/(2
\epsilon) -1\}$. We
divide our range and conclude for each part separately; since $\epsilon < 1/8$, we have that $\tan \epsilon \pi < 1$. Then, $\cos
\alpha - \tan (\pi \epsilon)  \sin \alpha\geq -1$ in the range $\alpha \in
[0,\pi/2]$. For $\alpha  \in [\pi/2,\pi-\pi
\epsilon]$, it can be shown that the function $\cos \alpha - \tan (\pi \epsilon) \sin \alpha$
is monotonically decreasing, thus it suffices to show that the inequality holds for
$n=1/(2\epsilon) -1$. Indeed,  
\begin{align*}
\cos (\pi -2 \epsilon \pi) - \tan (\epsilon \pi) \sin(\pi -2 \epsilon
\pi) & = -\cos(2 \pi \epsilon) - 2 \sin^2(\pi \epsilon) \\ & = 
-\cos^2(\pi \epsilon)-\sin^2(\pi \epsilon) \\ & = -1~.
\end{align*}
Therefore, we obtain from \eqref{eq:geq1} and \eqref{eq:geq2} that
\begin{align*}
\frac{1}{\sqrt{2}}\inner{w^* ,a_n }& \geq \frac{1}{\sqrt{2}} (-w^*[1] +
 w^*[3]) = \inner{w^*,(-1/\sqrt{2}, 0, 1/\sqrt{2})} = \inner{w^*,a_{1/2\epsilon}}  >  0,
\end{align*}
where the last inequality follows from the assumption that $h^*(a_{1/2\epsilon})=1$.

\end{proof}
%%% Local Variables:
%%% mode: latex
%%% TeX-master: "ALuMA"
%%% End:

\end{document}

%% file: header.tex
\usepackage{hyperref}

\usepackage{amsmath}

\usepackage{algorithm}
\usepackage{algorithmic}
\algsetup{indent=2em}

\newtheorem{ex}[theorem]{Example}

\newcommand{\sgn}{{\mathrm{sgn}}}

\newcommand{\reals}{\mathbb{R}}

\newcommand{\cH}{\mathcal{H}}
\renewcommand{\H}{\mathcal{H}}
\newcommand{\cL}{\mathcal{L}}  
\newcommand{\E}{\mathbb{E}}

\newcommand{\W}{\mathcal{W}}
\newcommand{\inner}[1]{\langle #1 \rangle}
\newcommand{\half}{\frac{1}{2}}

\newcommand{\eqdef}{\stackrel{\mathrm{def}}{=}}

\newcommand{\opt}{\mathrm{opt}}

\DeclareMathOperator*{\argmin}{argmin} % Declares argmin and ensures super and subscripts are located nicely above/below such as in \min
\DeclareMathOperator*{\argmax}{argmax} % Declares argmin and ensures super and subscripts are located nicely above/below such as in \min
\DeclareMathOperator*{\prob}{\mathbb{P}}

\renewcommand{\eqref}[1]{Equation~(\ref{#1})}
\newcommand{\exref}[1]{Example~(\ref{#1})}
\newcommand{\figref}[1]{Figure~\ref{#1}}
\newcommand{\secref}[1]{Section~\ref{#1}}
\newcommand{\appref}[1]{Appendix~\ref{#1}}
\newcommand{\thmref}[1]{Theorem~\ref{#1}}
\newcommand{\lemref}[1]{Lemma~\ref{#1}}

\newcommand{\corref}[1]{Corollary~\ref{#1}}